\documentclass[sigconf]{acmart}

\usepackage{todonotes}
\usepackage{booktabs}
\AtBeginDocument{%
  \providecommand\BibTeX{{%
    \normalfont B\kern-0.5em{\scshape i\kern-0.25em b}\kern-0.8em\TeX}}}

\usepackage{graphicx} 
\usepackage{bm} 
\usepackage{multirow}
\usepackage{amsmath}
\usepackage{mathtools}
\usepackage{graphicx}
\usepackage{subcaption}
\usepackage{enumitem}
\usepackage[version=4]{mhchem}

\begin{document}

\settopmatter{printacmref=false} 
\renewcommand\footnotetextcopyrightpermission[1]{} 
\pagestyle{plain} 

\title{{CHILI}: \underline{Ch}emically-\underline{I}nformed \underline{L}arge-scale \underline{I}norganic Nanomaterials Dataset for Advancing Graph Machine Learning}

\author{Ulrik Friis-Jensen}
\authornote{Both authors contributed equally to this research.}
\affiliation{%
  \institution{University of Copenhagen}
  \city{Copenhagen}
  \country{Denmark}
}
\email{ufj@chem.ku.dk}

\author{Frederik L. Johansen}
\authornotemark[1]
\affiliation{%
  \institution{University of Copenhagen}
  \city{Copenhagen}
  \country{Denmark}
}
\email{frjo@di.ku.dk}

\author{Andy S. Anker}
\affiliation{%
  \institution{Technical University of Denmark, DK}
  \country{}
}
\affiliation{%
  \institution{University of Oxford, UK}
  \country{}
}
\email{ansoan@dtu.dk}

\author{Erik B. Dam}
\affiliation{%
  \institution{University of Copenhagen}
  \city{Copenhagen}
  \country{Denmark}
}
\email{erikdam@di.ku.dk}

\author{Kirsten M. Ø. Jensen}
\affiliation{%
  \institution{University of Copenhagen}
  \city{Copenhagen}
  \country{Denmark}
}
\email{kirsten@chem.ku.dk}

\author{Raghavendra Selvan}
\affiliation{%
  \institution{University of Copenhagen}
  \city{Copenhagen}
  \country{Denmark}
}
\email{raghav@di.ku.dk}

\renewcommand{\shortauthors}{Friis-Jensen and Johansen, et al.}

\begin{abstract}
Advances in graph machine learning (ML) have been driven by applications in chemistry as graphs have remained the most expressive representations of molecules. 
This has led to progress within both fields, as challenging chemical data has helped improve existing methods and to develop new ones.
While early graph ML methods focused primarily on small organic molecules, more recently, the scope of graph ML has expanded to include inorganic materials. Modelling the periodicity and symmetry of inorganic crystalline materials poses unique challenges, which existing graph ML methods are unable to immediately address. Moving to inorganic nanomaterials further increases complexity as the scale of number of nodes within each graph can be broad ($10$ to $10^5$).

In addition, the bulk of existing graph ML focuses on characterising molecules and materials by predicting target properties with graphs as input. The most exciting applications of graph ML will be in their generative capabilities, in order to explore the vast chemical space from a data-driven perspective. Currently, generative modelling of graphs is not at par with other domains such as images or text, as generating chemically valid molecules and materials of varying properties is not straightforward. 

In this work, we invite the graph ML community to address these open challenges by presenting two new chemically-informed large-scale inorganic ({\tt CHILI}) nanomaterials datasets.  These datasets contain nanomaterials of different scales and properties represented as graphs of varying sizes. The first dataset is a medium-scale dataset (with overall >6M nodes, >49M edges) of mono-metallic oxide nanomaterials generated from 12 selected crystal types ({\tt CHILI-3K}). This dataset has a narrower chemical scope focused on an interesting part of chemical space with a lot of active research. The second is a large-scale dataset (with overall >183M nodes, >1.2B edges) of nanomaterials generated from experimentally determined crystal structures ({\tt CHILI-100K}). The crystal structures used in {\tt CHILI-100K} are obtained from a curated subset from the Crystallography Open Database (COD) and has a broader chemical scope covering database entries for 68 metals and 11 non-metals. We define 11 property prediction tasks covering node-, edge-, and graph- level tasks that span classification and regression. In addition we also define structure prediction tasks, which are of special interest for nanomaterial research. We benchmark the performance of a wide array of baseline methods starting with simple baselines to multiple off-the-shelf graph neural networks. 
Based on these benchmarking results, we highlight areas which need future work to achieve useful performance for applications in (nano)materials chemistry. To the best of our knowledge, {\tt CHILI-3K} and {\tt CHILI-100K} are the first open-source nanomaterial datasets of this scale -- both on the individual graph level and of the dataset as a whole -- and the only nanomaterials datasets with high structural and elemental diversity.\footnote{The datasets and benchmarking scripts are open-source, and available at \url{https://github.com/UlrikFriisJensen/CHILI}}
\end{abstract}


\maketitle
\newpage

\section{Introduction}
\label{sec:intro}
Graph machine learning (ML) has been applied to chemistry for more than 50 years~\cite{Jurs1971Machine}, and describing molecules as chemical graphs goes back as far as 1874~\cite{Cayley1874LVII}. A big part of why tasks in chemistry are so relevant for graph ML is that chemical structures, such as molecules, can be expressively represented using graphs, where atoms are encoded as nodes with local attributes and chemical bonds as edges with neighbourhood attributes~\cite{Hamilton2020Graph}. 

The recent advancements in deep learning methods~\cite{LeCun2015Deep,Schmidhuber2015Deep} have also impacted graph ML. The class of deep learning methods that have evolved to tackle non-Euclidean data can be viewed within the geometric deep learning paradigm~\cite{Scarselli2009Graph,Bronstein2017Geometric}. Specifically, the class of graph convolutional networks (GCNs) and graph neural networks (GNNs) have shown widespread applications on graph-structured data~\cite{Kipf2016Semi,Hu2020Open}. Several key developments in graph deep learning were specifically driven by applications on chemical graphs~\cite{Merkwirth2005Automatic,Duvenaud2015Convolutional, Gilmer2017Neural}. 

The primary focus of the graph ML community has been in modelling organic molecules~\cite{Wu2017MoleculeNet, Wang2023Graph}. Numerous studies have focused on predicting target properties of such molecules based on their atomic structure, posited as node- or graph- level tasks~\cite{Chen2019Graph, Bouritsas2023Improving, Hussain2022Global, Gilmer2017Neural, Schutt2017SchNet, Batatia2022MACE}. The inverse task of obtaining structures corresponding to desired properties has been studied to a lesser extent~\cite{DeCao2018MolGAN, Liu2021GraphEBM}. This is due to the inherent challenges of performing generative modelling in the space of graphs. While recent classes of methods such as denoising diffusion probabilistic models~\cite{Ho2020Denoising} are showing potential for smaller molecules~\cite{vignac2023digress,Zeni2023MatterGen}, the task of scaling these models to larger, and diverse, molecular graphs remains an open challenge.

The drive to develop novel materials with applications in batteries or catalysts for renewable energy storage has lead to new interest in inorganic crystalline materials, i.e materials with periodic atomic order~\cite{Hu2020Open, Tran2023Open}. A special case of such materials are nanocrystals, where the crystal dimensions are on the nanoscale. Modelling such materials pose new and interesting challenges that are unlike those encountered in organic chemistry applications~\cite{Anker2020Characterising}. Capturing the periodicity and symmetry of crystalline materials are not easily dealt with using existing graph ML methods~\cite{Xie2018Crystal, Chen2019Graph, Gong2022Examining, Cheng2021Geometric, Das2023CrysGNN}. This gets even more complex for nanomaterials where the scale of obtained graphs span a broad range of atoms, between  $10$ to $10^5$~\cite{Brus1995Electronic}.

In this paper, we invite the graph ML community to bridge the divide between existing methods to meet the complexity of large-scale inorganic materials chemistry. To this end, we make the following contributions:
\begin{enumerate}[leftmargin=*]
    \item Present a chemically-informed approach to generate large-scale graph datasets of nanomaterials. 
    \item Provide two novel nanomaterial graph datasets. 
    \item Outline an array of property- and structure- prediction tasks. 
    \item Perform comprehensive benchmark experiments for the proposed tasks.
\end{enumerate}

\begin{figure}[t]
    \centering
    \includegraphics[width=0.9\columnwidth]{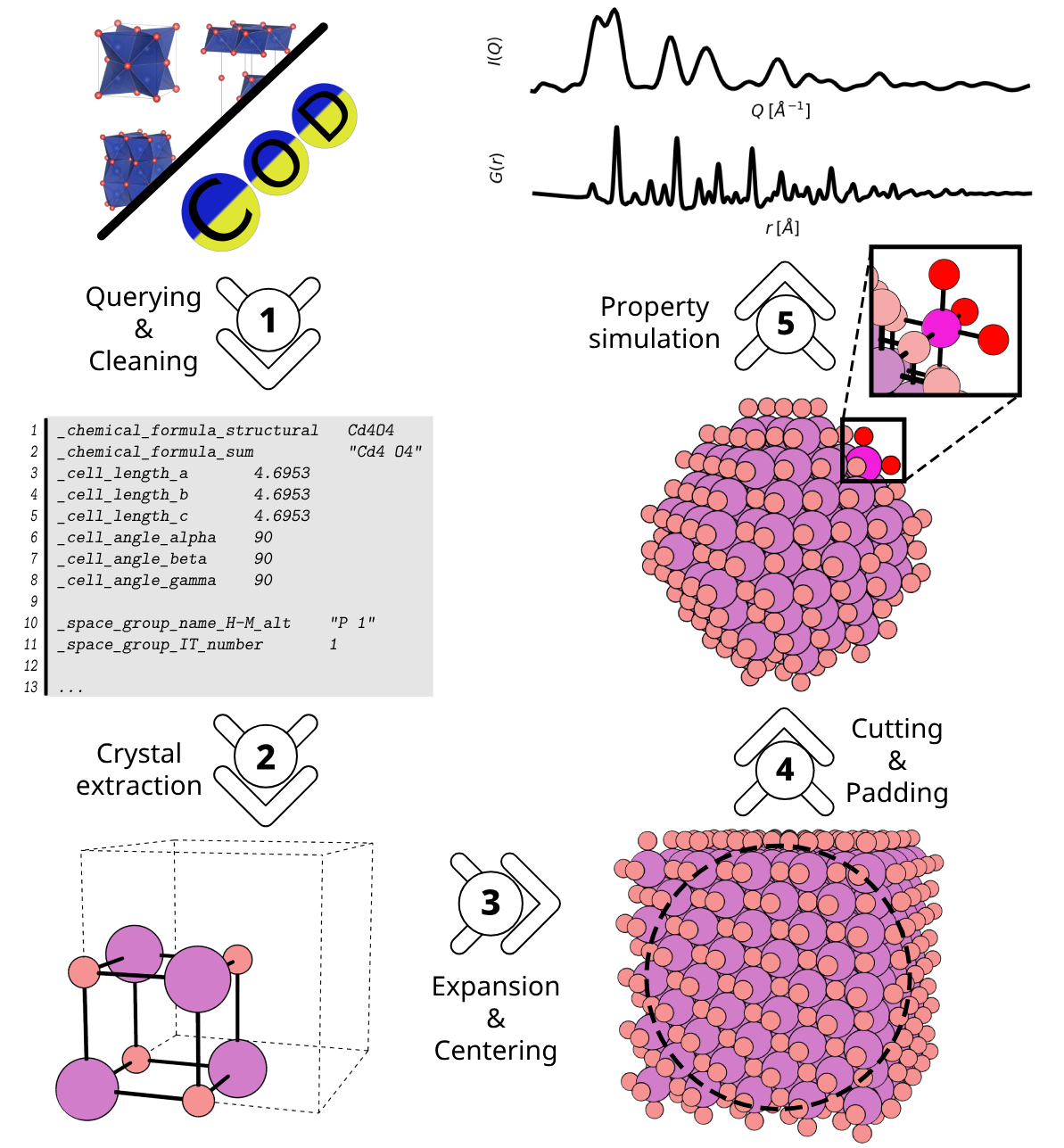}
    \caption{High-level schematic showing the five stages involved in the creation of the \texttt{CHILI}-datasets: (1) Querying and cleaning CIFs. (2) Extraction of crystal unit cells. (3) Expansion of unit cells into supercells and subsequent centering. (4) Cutting of nanoparticles into different sizes and padding of edge environments following the described rules, conversion into graphs with node- and edge- features. (5) Generation of graph-level properties from CIF (crystal type, crystal system, spacegroup, etc.) and simulation of scattering data.
    }
    \label{fig:generation_overview}
    \vspace{-0.5cm}
\end{figure}

An overview of our data-generating pipeline is illustrated in Figure~\ref{fig:generation_overview}. The data generation pipeline takes any source of crystallographic information files (CIFs) (a file format containing symmetry and positional information) and creates graphs of finite nanoparticles with variable sizes. We thus assume the structure of nanoparticles to be the cutouts of larger crystals, whether constructed manually or from a structural database. This is an approximation, as we do not take into account e.g. surface relaxation and size-dependent defects in the atomic structure. Each graph gets assigned relevant chemical labels, as well as simulated scattering data associated with each nanoparticle. The scattering data is simulated using DebyeCalculator~\cite{Johansen2024GPU}. Using this generation pipeline we provide two nanomaterial graph datasets, \texttt{CHILI-3K} and \texttt{CHILI-100K}. 

The {\tt CHILI-3K} dataset consists of $3180$ graphs representing mono-metallic oxide nanomaterials generated from $12$ different well known crystal structures, which are known to be taken by numerous materials. The resulting dataset captures a narrow chemical subspace that is of considerable interest due to their environmental, medical and catalytic applications~\cite{Ganachari2019Metal, Laurent2018Metal, Danish2020Systematic}. The second dataset, {\tt CHILI-100K}, consists of $104,408$ graphs generated by curating experimentally determined structures from the Crystallography Open Database (COD)~\cite{Graulis2009Crystallography}. This dataset has a broader chemical scope as it contains datapoints for materials consisting of combinations of 68 metals and 11 non-metals, thus spanning a much wider range of different crystal structures. While the {\tt CHILI-3K} is classified as a medium-scale dataset (with >6M nodes, >49M edges in total), the {\tt CHILI-100K} dataset is classified as a large-scale dataset (with >183M nodes, >1.2B edges in total) according to the Open Graph Benchmark (OGB) criteria~\cite{Hu2020Open}. The two {\tt CHILI} datasets are, to the best of our knowledge, the first open-source nanomaterial datasets of this scale. Together, we hope these two datasets will foster novel methodological contributions at the intersection of graph ML and large-scale inorganic materials chemistry.

\begin{table*}[t]
    \centering
    \caption{Overview of key summary statistics for the two proposed \texttt{CHILI} datasets. (*) indicates the total number of edges; two for each unique atom-atom pair.}
    \label{tab:dataset_summary}
    \vspace{-0.25cm}
    \footnotesize
    \begin{tabular}{@{\extracolsep{2pt}}lcccccccccccccc@{}}
    \toprule
        \multirow{2}{1em}{Dataset} & \multicolumn{4}{c}{\# of graphs}          & \multicolumn{4}{c}{\# of nodes}             & \multicolumn{4}{c}{\# of edges*}        & \multicolumn{2}{c}{Generation time}       \\
                                   \cmidrule{2-5}                                  \cmidrule{6-9}                                  \cmidrule{10-13}                            \cmidrule{14-15}
                                   & Total   & Train  & Validation  & Test      & Min & Median & Max    & Total              & Min & Median & Max     & Total             & Total            & Mean \\
    \midrule
        \texttt{CHILI-3K}                  & 3,180   & 2,544  & 318         & 318       & 7   & 1,377  & 14,793 & 6,959,085          & 7   & 7,212  & 118,258 & 49,624,440    & 02h51m29s         & 16.18s                 \\
        \texttt{CHILI-100K}                  & 104,408 & 83,526 & 10,441      & 10,441    & 2   & 1,054  & 21,427 & 183,398,463        & 2   & 5,336  & 413,762 & 1,251,841,365 & 68h25m12s         & 11.80s                 \\
    \bottomrule
    \end{tabular}
\end{table*}

\section{\texttt{CHILI} Datasets}
The \texttt{CHILI} datasets are generated following the approach shown in Figure~\ref{fig:generation_overview} as outlined in Section~\ref{sec:intro}. For a more detailed description of the graph generation see Appendix~\ref{app:nanoMaterialGraph} and~\ref{app:scatteringSimulation}. The summary statistics of the datasets are shown in Table~\ref{tab:dataset_summary} and the content of the datasets are described in the sections below.

\subsection{\texttt{CHILI-3K}}
The \texttt{CHILI-3K} dataset contains nanomaterial graphs generated from mono-metal oxides, which are a class of inorganic materials with a single metallic element coordinated to oxygen atoms in the structure. They are often studied for their many interesting applications~\cite{Ganachari2019Metal, Laurent2018Metal, Danish2020Systematic}.
The unit cells in \texttt{CHILI-3K} are constructed based on 12 of the crystal types described in West et al.~\cite{West2022Solid}, which are known to be formed by mono-metal oxides. The unit cells of the 12 crystal types are visualized in Figure~\ref{fig:crystal_types}. The CIF construction is described in Appendix~\ref{app:cif_gen}. 

The \texttt{CHILI-3K} dataset contains 53 metallic elements and only 1 non-metallic element, oxygen.
 All structures contain oxygen and one metal chosen from the 53 options as coloured blue in the periodic table in Figure \ref{fig:periodic_table}. Every combination of crystal type and metal is used, indiscriminately to whether the specific element is stable in the particular crystal types. This is done with the intention for the \texttt{CHILI-3K} dataset to achieve complete coverage of the well defined points in this chemical subspace, leading to a total of 636 unique CIFs. 
\begin{figure}[t]
    \centering
    \includegraphics[width=\columnwidth]{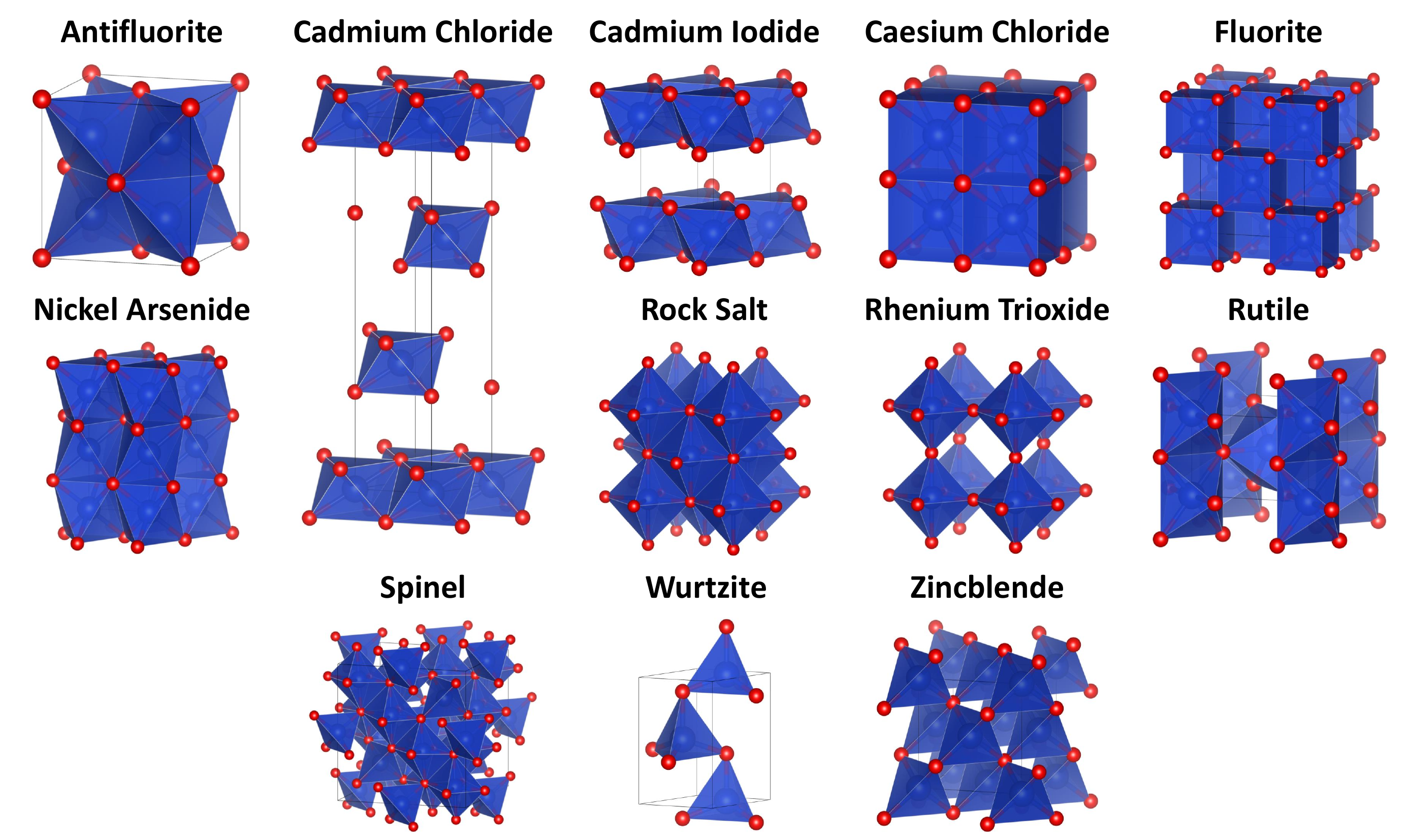}
    \caption{The unit cells of the 12 crystal types present in the \texttt{CHILI-3K} dataset. For all shown structures, copper (Cu) is the metal. The unit cells are visualized using VESTA~\cite{Momma2008VESTA} with the polyhedral style. The unit cells are shown from the standard orientation of a crystal shape, which is one of the 7 view options in VESTA.}
    \label{fig:crystal_types}
    \vspace{-0.35cm}
\end{figure}

\subsection{\texttt{CHILI-100K}}
\begin{figure}[t]
    \centering
    \includegraphics[width=0.48\textwidth]{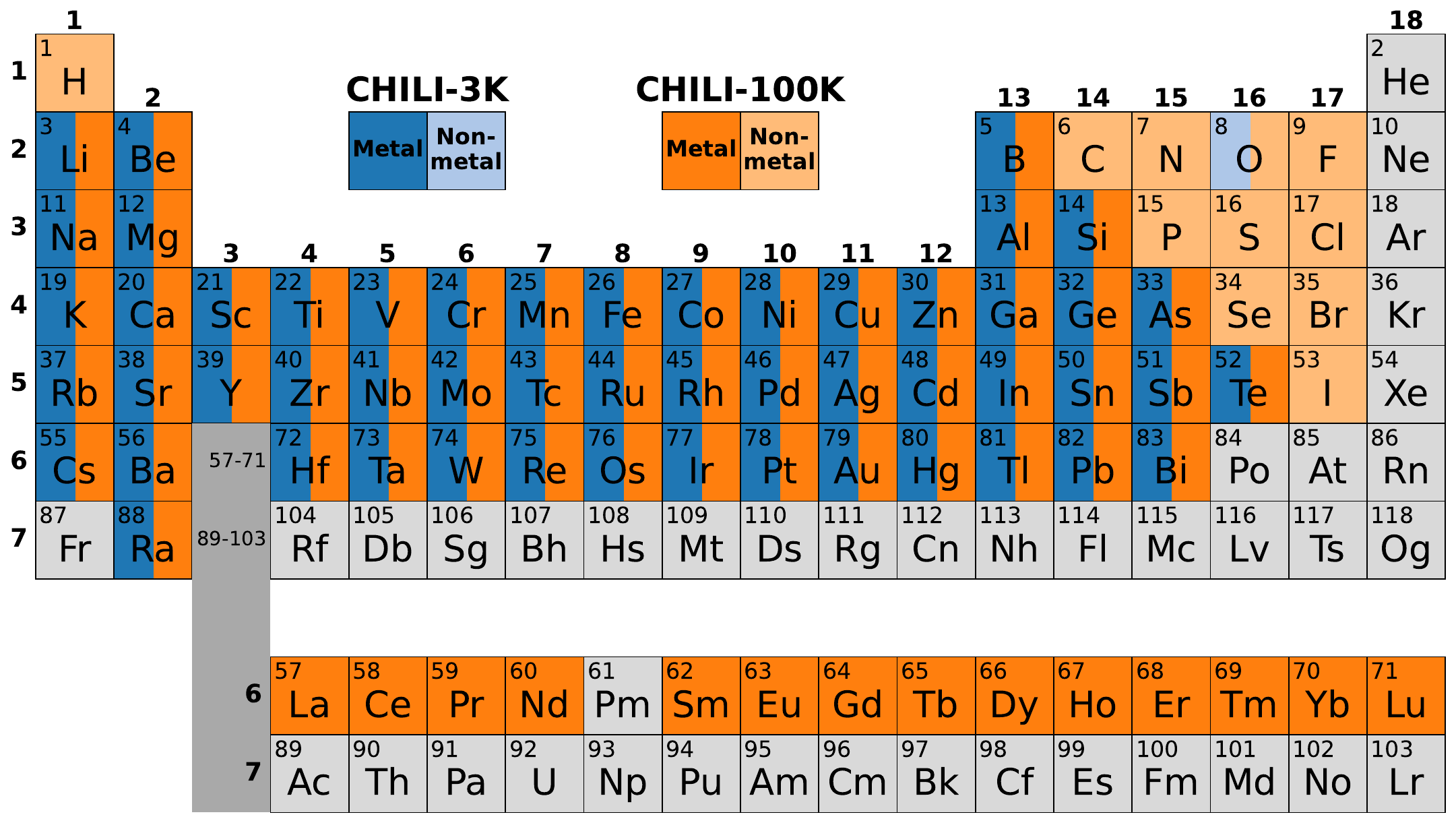}
    \caption{The periodic table with each element colored depending on if they are included in \texttt{CHILI-3K} (blue), \texttt{CHILI-100K} (orange) or none of them (light grey). The shade of the colors indicate whether the element is considered a metal (bright) or a non-metal (muted).}
    \label{fig:periodic_table}
    \vspace{-0.5cm}
\end{figure}

The \texttt{CHILI-100K} dataset consists of nanomaterial graphs generated from a subset of inorganic materials from COD~\cite{Graulis2009Crystallography}. The subset includes materials which contains any of the 68 metals and 11 non-metals shown in orange in Figure~\ref{fig:periodic_table}. COD was queried for both purely metallic phases and phases including combinations metals and non-metals, like metal oxides. Materials containing elements not included in the shown selection were removed. For simplicity, only materials with unit-cell volumes smaller than $1,000$ Å$^3$ were used in order to avoid the inclusion of e.g. metal organic frameworks and larger inorganic coordination complexes. The result of the COD query was saved as COD IDs in a csv file, which can be found in the GitHub repository\footnote{\scriptsize\url{https://github.com/UlrikFriisJensen/CHILI/blob/main/generation/COD_subset_IDs.csv}}. 

The corresponding, downloaded CIFs were cleaned for issues that caused the CIF to be unreadable or could potentially impact the further analysis. From the original query to COD we obtain $61794$ COD IDs, which reduces to $20882$ usable CIFs after cleaning. The specific choices used to clean COD to obtain {\tt CHILI-100K} are reported in Appendix~\ref{app:chili100k}.

The \texttt{CHILI-100K} dataset is intended to mimic the data distribution of real world materials. However, we want to emphasize that there is an inherent bias with using a database comprising experimental materials. This bias is towards known materials, which are stable and easier to synthesize, but it does not cover all possible materials in the chosen chemical subspace. This is especially important to consider when tasked with generating novel materials using \texttt{CHILI-100K}.

\begin{table*}[t]
    \centering
    \caption{Overview of the \texttt{CHILI} data structure with different variables. Depending on the nature of the task, these attributes can be used as input or as target variables.}
    \vspace{-0.25cm}
    \label{tab:data_structure}
    \footnotesize
    \begin{tabular}{cccl}
        \toprule
         \multicolumn{2}{c}{Name} & \multirow{2}*{Shape} & \multirow{2}*{Description}\\
         \cline{1-2}
         Level 1 & Level 2 &&\\
         \midrule
         {\tt x}& & [n\_atoms, 4] & Node feature matrix. [\texttt{atomic\_number}, \texttt{atomic\_radius}, \texttt{atomic\_weight}, \texttt{electron\_affinity}]\\
         {\tt edge\_index}& & [2, n\_bonds]& Graph connectivity in sparse coordinate (COO) format.\\
         {\tt edge\_attr}& & [n\_bonds, 1]& Edge feature matrix. [distance (Å)]\\
         \multirow{23}{1em}{\tt y}&{\tt crystal\_type}& ---& Name of the crystal type or "Unknown".\\
         & {\tt space\_group\_symbol}& ---&Space group symbol of the unit cell in Hermann-Mauguinn/international notation.\\
         &{\tt space\_group\_number}& ---& Space group number of the unit cell.\\
         &{\tt crystal\_system}& ---& Name of the crystal system.\\
         &{\tt crystal\_system\_number}& ---& Logit corresponding to the crystal system from lowest to highest symmetry.\\
         &{\tt atomic\_species}& [n\_atomic\_species]& Unique atom number matrix.\\
         &{\tt n\_atomic\_species}& ---& Number of unique atomic species in the nanoparticle.\\
         &{\tt np\_size}& ---& Diameter of the nanoparticle measured in Ångströms (1 Å = $10^{-10}$ m).\\
         &{\tt n\_atoms}& ---& Number of atoms in the nanoparticle.\\
         &{\tt n\_bonds}& ---& Number of ``bonds" in the nanoparticle.\\
         &{\tt cell\_params}& [6]&Cell parameter matrix. The cell parameters are [a, b, c, $\alpha$, $\beta$, $\gamma$].\\
         &{\tt unit\_cell\_node\_feat}& [n\_atoms\_unit\_cell, 4] &Node feature matrix for the unit cell.\\
         &{\tt unit\_cell\_edge\_index}& [2, n\_bonds\_unit\_cell]&Graph connectivity in COO format.\\
          &{\tt unit\_cell\_edge\_feat}& [n\_bonds\_unit\_cell, 1]&Edge feature matrix.\\
         &{\tt unit\_cell\_pos\_abs}& [n\_atoms\_unit\_cell, 3]&Node position matrix (absolute coordinates) for the unit cell.\\
         &{\tt unit\_cell\_pos\_frac}& [n\_atoms\_unit\_cell, 3]&Node position matrix (fractional coordinates) for the unit cell.\\
         &{\tt unit\_cell\_n\_atoms}& [n\_atoms\_unit\_cell]&Number of atoms in the unit cell.\\
         &{\tt unit\_cell\_n\_bonds}& [n\_bonds\_unit\_cell]&Number of ``bonds" in the unit cell.\\
         &{\tt nd}& [2, 580]&Simulated neutron diffraction (ND) from the nanoparticle.\\
         &{\tt xrd}& [2, 580]&Simulated X-ray diffraction (XRD) from the nanoparticle.\\
         &{\tt nPDF}& [2, 6000]&Simulated neutron pair distribution function (PDF) from the nanoparticle.\\
         &{\tt xPDF}& [2, 6000]&Simulated X-ray pair distribution function (PDF) from the nanoparticle.\\
         &{\tt sans}& [2, 300]&Simulated small-angle neutron scattering (SANS) from the nanoparticle.\\
         &{\tt saxs}& [2, 300]&Simulated small-angle X-ray scattering (SAXS) from the nanoparticle.\\
         {\tt pos\_frac}& & [n\_atoms, 3]& Node position matrix (fractional coordinates).\\
         {\tt pos\_abs}& & [n\_atoms, 3]& Node position matrix (absolute coordinates).\\
         \bottomrule
    \end{tabular}
    \vspace{-0.25cm}
\end{table*}

\begin{figure*}[t]
  \begin{subfigure}[t]{0.32\textwidth}
    \centering
    \includegraphics[width=\columnwidth]{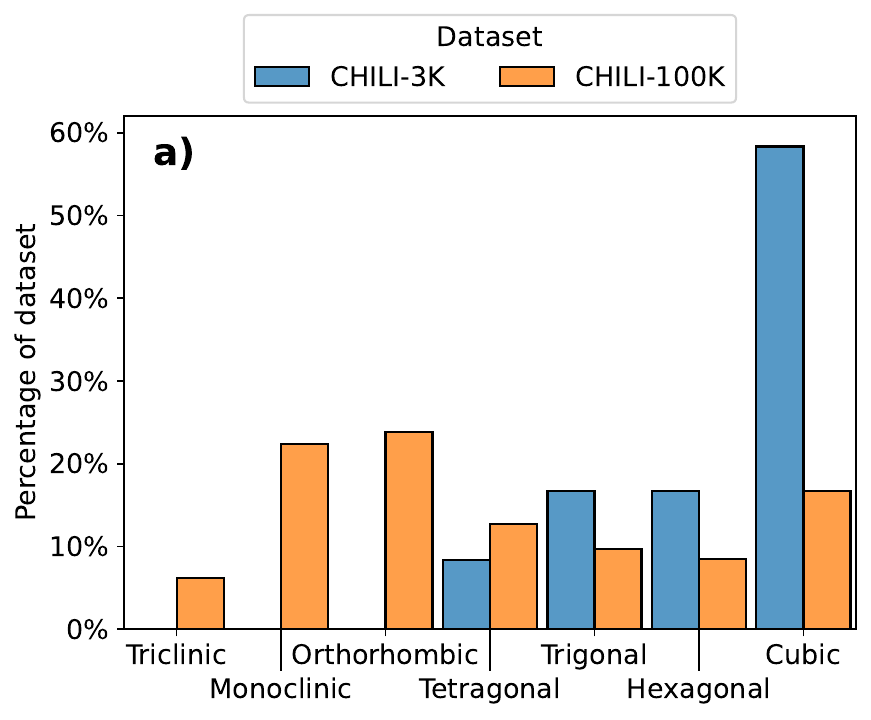}
    \vspace{-0.35cm}
    \phantomcaption{}
    \label{fig:dist_crystal_system}
    \end{subfigure}
\begin{subfigure}[t]{0.32\textwidth}
    \centering
    \includegraphics[width=\columnwidth]{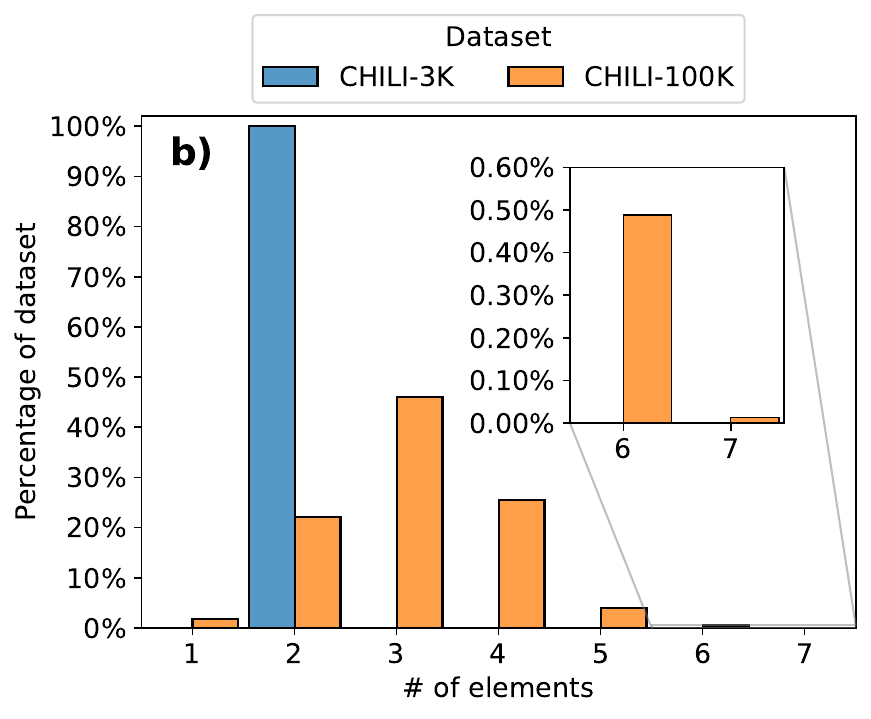}
    \vspace{-0.35cm}
    \phantomcaption{}
    \label{fig:dist_num_elements}
\end{subfigure}
\begin{subfigure}[t]{0.32\textwidth}
    \centering
    \includegraphics[width=\columnwidth]{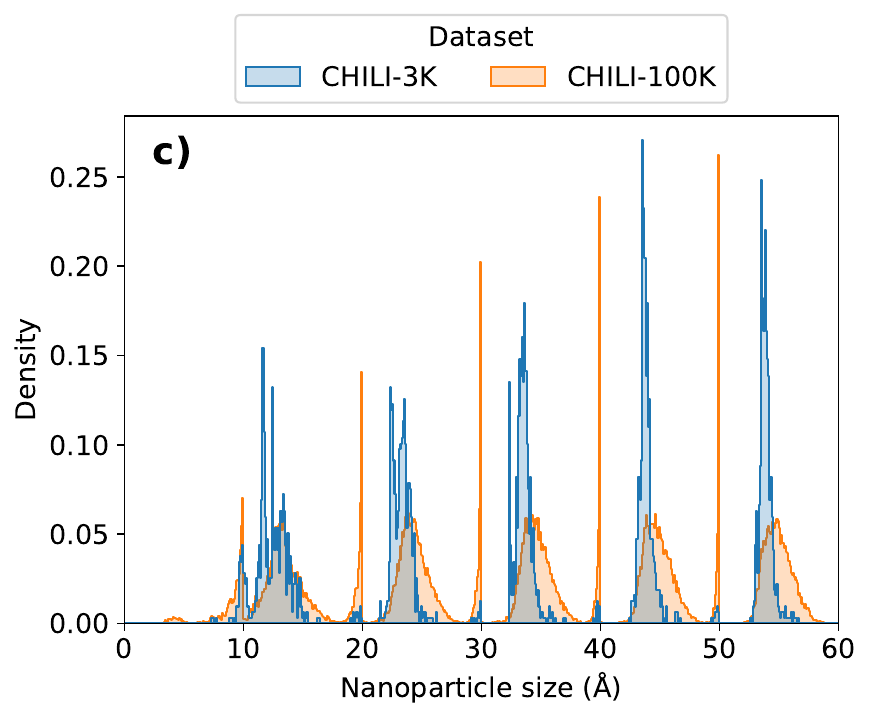}
    \vspace{-0.35cm}
    \phantomcaption{}
    \label{fig:dist_sizes}
    \end{subfigure}
    \vspace{-0.3cm}
    \caption{a) Distribution of crystal systems in the \texttt{CHILI-3K} dataset (blue) and the \texttt{CHILI-100K} dataset (orange). b) Distribution of the number of unique elements in each structure for the \texttt{CHILI-3K} dataset (blue) and the \texttt{CHILI-100K} dataset (orange). The inset plot shows 6 and 7 elements at a more appropriate y-axis scale. c) Distribution of the size of the generated nanoparticles for the \texttt{CHILI-3K} dataset (blue) and the \texttt{CHILI-100K} dataset (orange).}
    \vspace{-0.25cm}
\end{figure*}

\subsection{Data structure}

The data objects follow standard graph data structure common in graph ML comprising node features as ${\tt x}$, edges specified in COO format as ${\tt edge\_index}$, edge features as ${\tt edge\_attr}$ and the graph labels stored in a dictionary ${\tt y}$. Taking inspiration from molecular graph datasets, like QM9~\cite{Ruddigkeit2012Enumeration, Ramakrishnan2014Quantum}, the atomic positions are not given with the other node features, but instead as separate attributes for the absolute atomic coordinates, ${\tt pos\_abs}$, and the fractional atomic coordinates, ${\tt pos\_frac}$. Different target properties that are of interest for inorganic chemistry are generated for both the datasets and included to be either used as target variables or conditioning inputs. A detailed overview of the data structure with all the input and target properties is presented in Table \ref{tab:data_structure}.

\subsection{Dataset statistics}
With the assumption that the true atomic structure of nanomaterials can be approximated using the unit cell of the crystalline material, it is meaningful to consider the distribution of crystal systems in the proposed datasets. The 7 crystal systems are a high level description of the symmetry present in the crystalline material, going from triclinic (lowest symmetry) to cubic (highest symmetry). The distribution of crystal systems in the \texttt{CHILI} datasets are shown in Figure \ref{fig:dist_crystal_system}. From this it is clear that \texttt{CHILI-3K} only has the 4 crystal systems with the highest symmetry present and that the cubic system is over-represented with almost 60 \% of the data. However, this imbalance matches the crystal systems of the crystal types, as defined in West et al.~\cite{West2022Solid}. The \texttt{CHILI-100K} dataset consists of all 7 crystal systems, but they are not equally distributed. The most represented crystal system is orthorhombic, with about 25 \% of the data, and the least represented crystal system is triclinic, with around 7 \% of the data. As the data was not selected from COD based on the crystal systems, we can assume that the real distribution of crystal systems in inorganic materials is roughly the same as we observe here.

It is also interesting to look at the distribution of number of unique elements in the nanomaterials in the two datasets, as this can be used as an additional way of determining structural diversity in the dataset. From Figure~\ref{fig:dist_num_elements} we see that \texttt{CHILI-3K} only contains nanomaterials with 2 elements, as it was constructed in this way. For \texttt{CHILI-100K} the number of elements range from 1 to 7, with most nanomaterials having between 2 and 4 elements.

As we are dealing with nanomaterials that have a finite size, it is also relevant to look at the distribution of sizes in the two datasets. Referring to Figure~\ref{fig:dist_sizes}, we observe that while some nanoparticles match the specified generation size precisely, most structures appear larger. This comes from the non-metals coordinated to the metallic particle core. Note as well, that the larger structural diversity in \texttt{CHILI-100K} constitutes to a smoother size distribution than that of \texttt{CHILI-3K}.
Additional dataset statistics are provided in Appendix \ref{app:statistics}. 

\section{Related Work}
\label{sec:related}
The intersection of chemistry and graph ML is an active area of research that has resulted in several interesting data sources, methods, and research directions. A short review of these are presented in this section.

\subsection{Data sources}
\textbf{Graph datasets:}
Graph datasets can vary a lot depending on the domain the graphs are from, their scales, and the types of tasks they are designed to be used for. The OGB benchmark contains a broad overview of existing graph datasets~\cite{Hu2020Open}, with single-  and multi- graph datasets being the two high-level types. The former consists of a single large graph, like knowledge graphs~\cite{Thanapalasingam2023IntelliGraphs} and social networks~\cite{Leskovec2012Learning}, which then has nodes and edges split into training, validation and test sets. These datasets are primarily intended for transductive learning such as node- or edge-level prediction tasks~\cite{Kipf2016Semi}. Multi-graph datasets consists of many, often small, graphs. These could be molecular graphs~\cite{Ruddigkeit2012Enumeration} or protein-protein interaction graphs~\cite{Zitnik2017Predicting}, which are then split into training, validation and test set at the graph level. These kinds of datasets are mostly used for inductive learning at the graph-level, but both node- and edge-level tasks are feasible too~\cite{Hamilton2017Inductive}.

\textbf{Molecular graph datasets:} As molecules can be easily represented as graphs, datasets consisting of molecular graphs have played an important part in the development of GNNs~\cite{Duvenaud2015Convolutional,Gilmer2017Neural}. A wide variety of molecular graph datasets exists for many different purposes, as seen in QM9~\cite{Ruddigkeit2012Enumeration, Ramakrishnan2014Quantum} which is focused on density functional theory (DFT) calculated molecular properties, ZINC~\cite{Sterling2015ZINC}, which is focused on drug discovery, and MoleculeNet~\cite{Wu2017MoleculeNet}, which is a benchmark collection of molecular graph datasets within quantum chemistry, physical chemistry, biophysics and physiology.

\textbf{Structural databases:}
The structural databases in chemistry mainly contain crystal structures solved from crystallography experiments, and in some cases also synthetic crystal structures, which are stored in CIFs~\cite{Brown1996CIF}. A wide variety of structural databases exists today, with some of the most widely used being the Cambridge Structural Database (CSD)~\cite{Groom2016Cambridge}, the Inorganic Crystallography Structure Database (ICSD)~\cite{Zagorac2019Recent}, the COD~\cite{Downs2003TheAM, Graulis2009Crystallography, Graulis2011Crystallography, Graulis2015Computing, Merkys2016COD, Quirs2018Using, Vaitkus2021Validation, Merkys2023Graph, Vaitkus2023workflow} and the Materials Project (MP)~\cite{Jain2013Commentary}. 

\textbf{Material graph datasets:}
The representation of materials are similar to molecules in many ways, but differ in two critical aspects. Firstly, unlike molecules, materials feature diverse bonding types across varying scales. This makes the translation of the atoms in a material into nodes and edges less trivial. Secondly, materials are periodic and often described only as their smallest repeating unit -- the unit cell -- and a set of symmetry operations called the spacegroup. Material graphs are, therefore, of similar scale as molecular graphs, but with added periodicity through the edges. Recently material graphs have garnered more interest with the growing popularity of MP~\cite{Jain2013Commentary}, the publication of the Open Quantum Materials Database (OQMD)~\cite{Kirklin2015Open} and the Open Catalyst datasets: OC20~\cite{Chanussot2021Open} and OC22~\cite{Tran2023Open}. 
The primary focus with these datasets has been in the discovery of new catalysts and energy storage materials~\cite{Choudhary2023Large,Lee2023MatSciML,Zeni2023MatterGen}, approximating DFT with ML~\cite{Jung2023High} and most recently foundational models for materials~\cite{Batatia2024Foundation}.

\textbf{Nanomaterial graph datasets:}
Nanomaterials are distinguished from other materials by their finite and small size, which invalidates the assumption of long-range periodicity. Thus, to properly describe a nanomaterial, all atoms must be explicitly defined, which significantly increases the scale of a nanomaterial graphs compared to molecules and materials. 
Nanomaterials can be further categorized into nanoclusters -- nanomaterials with a size of < 1nm and up to 100-150 atoms -- and nanoparticles -- nanomaterials with a size between 1 and 100 nm and upwards of 10's - 100's of thousands of atoms -- ~\cite{Yang2022Big, Brus1995Electronic}.
To our knowledge, only a few nanomaterial graph datasets exist in the literature. We only know of one nanocluster graph dataset from Fung et al~\cite{fung2021benchmarking} (data from~\cite{Fung2017Exploring}) with 20,000 configurations of 10 to 13 platinum atoms. We also know of some non-graph datasets related to nanomaterials, like the dataset from Manna et al.~\cite{Manna2023Database} with 63,015 DFT relaxed nanoclusters from 55 different elements and the dataset from Barnard et al.~\cite{Barnard2017Silver} with 425 silver nanoparticles ranging in size from 13 to 2947 atoms. 

\subsection{Graph ML tasks}
\textbf{Material property prediction:} 
Predicting material properties using graph ML is an inductive learning task that necessitates generalization of learnt behaviour at node- and graph- levels to unseen chemical graphs~\cite{Hamilton2017Inductive}. There have been several attempts at feature-engineering as a means to do so; as this would circumvent the need for graph representations~\cite{Wang2018Dynamic, Jha2019IRNet, Jha2018ElemNet, Goodall2020Predicting}. These methods use descriptors from the materials that are agnostic to the arrangement of atoms within them. They rely solely on attributes such as stoichiometry, elemental statistics, electronic structure and the ionic compound. For instance, an ML framework for formation energy prediction using a total of 145 attributes including those mentioned above was proposed in Ward et al.~\cite{Ward2016general}. Other studies consider the periodic crystal structures of the materials and expand upon the principles of crystallography to derive periodic graphs which gives insights into material properties~\cite{Louis2020Graph, Schmidt2021Crystal, Xie2018Crystal, Choudhary2021Atomistic}. Some salient examples include the crystal graph convolution network~\cite{Xie2018Crystal}, that aims to capture atom interactions across the cell boundaries by using multi-edge graphs, and the atomistic line GNN~\cite{Choudhary2021Atomistic}, that considers not only the distances between neighboring atoms but also the angles at which they are arranged within the crystal lattice. These models have been shown to perform well on predicting several structure properties, including formation energies. Still, to our knowledge, only few attempts have been made to model properties associated with materials that lack long-range order; like nanomaterials, clusters, as well as amorphous- and disordered materials~\cite{fung2021benchmarking, Banik2023CEGANN}.

\textbf{Inverse materials design:} Using graph ML for materials design poses a more challenging task than property prediction tasks. In this setting, the structure, or a set of potential structures, of a material that match desired properties or functions are explored. Latent variable models are used to this end, as they can be used in generative settings while respecting any conditioning inputs like the atomic composition of the material, catalytic ability, stability or structural characteristics such as those derived from X-ray and neutron diffraction experiments~\cite{Anker2020Characterising}. Xie et al., for instance, employ a crystal-diffusion variational autoencoder to generate stable crystal structures~\cite{Xie2021Crystal}. They encode materials into a lower-dimensional latent space, from where a property predictor network predicts attributes such as composition and lattice parameters and translates them into crystal structures by applying them to randomly initialized unit cell structures. A GNN is then used to optimize the atomic positions until reaching equilibrium. In a similar manner, Merchant et al. investigates the use of GNNs for the purpose of discovering new crystal structures in their recent work~\cite{Merchant2023Scaling}. The iterative training process of these specific GNNs, known as Graph Networks for Materials Exploration, includes filtering potential structures according to initial predictions and then validating energy computations with DFT. When it comes to inverse design for nanomaterials on the other hand, little progress has been made. Recently however, some advancements have been achieved in determining the atomic structure of nanomaterials using scattering data. Specifically, Kjær et al. introduce DeepStruc~\cite{Kjaer2023DeepStruc}, a  conditional variational autoencoder that is capable of determining the atomic structures of a subset of mono-metallic nanoparticles directly from their pair distribution functions (PDFs) derived from total scattering data. DeepStruc successfully determines structures from PDFs originating from seven distinct structure types. This is achieved on both simulated and experimental PDFs.

\section{Experiments and benchmarking}

To characterise the two {\tt CHILI} datasets across different experiments we perform a selection of classification-, regression- and structure generation- tasks. For each task, we also report simpler baselines like random prediction, most frequent class (MFC), and mean prediction, wherever appropriate. We use a range of off-the-shelf GNN models as backbone networks and add an additional linear layer as the last layer to predict the appropriate number of classes or regression targets. The GNN backbone models range from simple to complex models: GCN~\cite{Kipf2016Semi}, PMLP~\cite{Yang2022Graph}, GraphSAGE~\cite{Hamilton2017Inductive}, GAT~\cite{velivckovic2018graph}, GraphUNet~\cite{Gao2019Graph}, GIN~\cite{Xu2018How}, and EdgeCNN~\cite{Wang2018Dynamic}. An overview of the prediction tasks and the corresponding output dimensions are shown in Table~\ref{tab:tasks_overview}.

\begin{table}[t]
    \caption{Overview of the proposed tasks.}
    \label{tab:tasks}
    \centering
    \vspace{-0.25cm}
    \footnotesize
    \begin{tabular}{lcc}
    \toprule
         Classification task                          & Level & \# of classes \\
        \midrule
         \texttt{atomic\_number}            & Node  & 118           \\
         \texttt{crystal\_system\_number}  & Graph & 7             \\
         \texttt{space\_group\_number}     & Graph & 230           \\
    \toprule
         Regression task                           & Level & Target size \\
         \midrule
         \texttt{pos\_abs} (position)             & Node  & 3           \\
         \texttt{edge\_attr} (distance)            & Edge  & 1           \\
         \texttt{saxs}                 & Graph & 300         \\
         \texttt{sans}                 & Graph & 300         \\
         \texttt{xrd}                  & Graph & 580         \\
         \texttt{nd}                   & Graph & 580         \\
         \texttt{xPDF}                 & Graph & 6000        \\
         \texttt{nPDF}                 & Graph & 6000        \\
    \toprule
         Structure generation task                                           & \multicolumn{2}{c}{max \# atoms} \\
         \midrule
         \texttt{saxs} $\rightarrow$ \texttt{unit\_cell\_pos\_frac}                 & \multicolumn{2}{c}{20}           \\
         \texttt{xrd} $\rightarrow$ \texttt{unit\_cell\_pos\_frac}                    & \multicolumn{2}{c}{20}           \\
         \texttt{xPDF} $\rightarrow$ \texttt{unit\_cell\_pos\_frac}                 & \multicolumn{2}{c}{20}           \\
          
         \texttt{saxs} $\rightarrow$ \texttt{pos\_abs}                 & \multicolumn{2}{c}{200}           \\
         \texttt{xrd} $\rightarrow$  \texttt{pos\_abs}                   & \multicolumn{2}{c}{200}           \\
         \texttt{xPDF} $\rightarrow$ \texttt{pos\_abs}                & \multicolumn{2}{c}{200}           \\
         \bottomrule
    \end{tabular}
    \label{tab:tasks_overview}
    \vspace{-1cm}
\end{table}

\begin{table*}[t]
    \caption{Overview of property prediction benchmark experiments on the \texttt{CHILI-3K} and \texttt{CHILI-100K} datasets. All experiments were repeated 3 times with different seeds, early stopping with a patience of 50 epochs and a maximum training time of 1 hour. Random: random class prediction, MFC: Most Frequent Class. Models are reported in increasing order of complexity measured in terms of the number of trainable parameters in the GNN backbone.}
    \label{tab:results_predictive}
    \footnotesize
    \begin{tabular}{@{\extracolsep{1pt}}llcccccccc@{}}
    \toprule
        &Task type  & \multicolumn{3}{c}{Classification}                 & \multicolumn{5}{c}{Regression}                                                                  \\
          \cmidrule{3-5} \cmidrule{6-10}
      &Task level    & Node   & \multicolumn{2}{c}{Graph}     & Node                 & Edge        & \multicolumn{3}{c}{Graph}                 \\
          \cmidrule{3-3} \cmidrule{4-5} \cmidrule{6-6} \cmidrule{7-7} \cmidrule{8-10}
      &Target  & \texttt{atomic\_num.}                   & \texttt{ crystal\_system\_num.}      & \texttt{ space\_group\_num.}          & \texttt{ pos\_abs}                    & \texttt{ edge\_attr}               & \texttt{ saxs}                & \texttt{ xrd}               & \texttt{ xPDF}                   \\
          \cmidrule{3-5} \cmidrule{6-6} \cmidrule{7-10}
      &Metric  & \multicolumn{3}{c}{Weighted F1-score ($\uparrow$)}                  & Pos. MAE ($\downarrow$) & \multicolumn{4}{c}{MSE ($\downarrow$)}                                                            \\
    \midrule
    
\multirow[c]{10}{*}{\rotatebox[origin=c]{90}{\texttt{CHILI-3K}}} & Random & $0.016 \pm$ \tiny{0.000} & $0.191 \pm$ \tiny{0.008} & $0.009 \pm$ \tiny{0.008} & --- & --- & --- & --- & --- \\
 & MFC & $0.461$ & $0.440$ & $0.108$ & --- & --- & --- & --- & --- \\
 & Mean & --- & --- & --- & $16.575$ & $0.265$ & $0.037$ & $0.017$ & $\bm{0.008}$ \\ \cmidrule{2-10}
 & GCN~\cite{Kipf2016Semi} & $0.496 \pm $ \tiny{0.001} & $0.367 \pm $ \tiny{0.127} & $0.099 \pm $ \tiny{0.019} & $16.575 \pm $ \tiny{0.000} & $0.056 \pm $ \tiny{0.006} & $0.008 \pm $ \tiny{0.000} & $0.010 \pm $ \tiny{0.000} & $0.012 \pm $ \tiny{0.000} \\
 & PMLP~\cite{Yang2022Graph} & $0.461 \pm $ \tiny{0.000} & $0.440 \pm $ \tiny{0.036} & $0.135 \pm $ \tiny{0.006} & $16.575 \pm $ \tiny{0.000} & $0.359 \pm $ \tiny{0.017} & $0.022 \pm $ \tiny{0.025} & $0.010 \pm $ \tiny{0.000} & $0.012 \pm $ \tiny{0.000} \\
 & GraphSAGE~\cite{Hamilton2017Inductive} & $0.491 \pm $ \tiny{0.004} & $0.422 \pm $ \tiny{0.037} & $0.151 \pm $ \tiny{0.045} & $16.575 \pm $ \tiny{0.000} & $0.055 \pm $ \tiny{0.002} & $0.008 \pm $ \tiny{0.001} & $0.010 \pm $ \tiny{0.000} & $0.012 \pm $ \tiny{0.000} \\
 & GAT~\cite{velivckovic2018graph} & $0.461 \pm $ \tiny{0.000} & $0.504 \pm $ \tiny{0.076} & $0.113 \pm $ \tiny{0.013} & $16.575 \pm $ \tiny{0.000} & $0.342 \pm $ \tiny{0.117} & $0.008 \pm $ \tiny{0.000} & $0.010 \pm $ \tiny{0.000} & $0.029 \pm $ \tiny{0.030} \\
 & GraphUNet~\cite{Gao2019Graph} & $0.552 \pm $ \tiny{0.079} & $0.431 \pm $ \tiny{0.014} & $0.095 \pm $ \tiny{0.036} & $\bm{14.765 \pm }$ \textbf{\tiny{0.395}} & $0.055 \pm $ \tiny{0.001} & $0.008 \pm $ \tiny{0.000} & $0.010 \pm $ \tiny{0.000} & $0.012 \pm $ \tiny{0.000} \\
 & GIN~\cite{Xu2018How} & $0.587 \pm $ \tiny{0.002} & $0.438 \pm $ \tiny{0.004} & $0.125 \pm $ \tiny{0.026} & $16.575 \pm $ \tiny{0.000} & $0.464 \pm $ \tiny{0.005} & $0.008 \pm $ \tiny{0.000} & Unstable & Unstable \\
 & EdgeCNN~\cite{Wang2018Dynamic} & $\bm{0.632 \pm }$ \textbf{\tiny{0.009}} & $\bm{0.657 \pm }$ \textbf{\tiny{0.196}} & $\bm{0.733 \pm }$ \textbf{\tiny{0.207}} & $16.575 \pm $ \tiny{0.000} & $\bm{0.015 \pm }$ \textbf{\tiny{0.001}} & $\bm{0.006 \pm }$ \textbf{\tiny{0.004}} & $\bm{0.008 \pm }$ \textbf{\tiny{0.001}} & $0.011 \pm $ \tiny{0.000} \\
\midrule
\midrule
\multirow[c]{10}{*}{\rotatebox[origin=c]{90}{\texttt{CHILI-100K}}} & Random & $ 0.015 \pm $ \tiny{0.000} & $\bm{0.168 \pm }$ \textbf{\tiny{0.014}} & $0.002 \pm $ \tiny{0.001} & --- & --- & --- & --- & --- \\
 & MFC                                      & $ 0.192 $                     & $ 0.046 $       & $ 0.010 $                      & ---                           & ---                  & ---                            & ---                            & ---              \\
 & Mean                                     & ---                           & ---             & ---                             & $ 16.336 $              & $ 0.307 $                  & $ 0.038 $                            & $ 0.021 $                            & $\bm{ 0.007 }$         \\
 \cmidrule{2-10}
 & GCN~\cite{Kipf2016Semi}     & $0.275 \pm $ \tiny{0.002} & $0.069 \pm $ \tiny{0.023} & $0.043 \pm $ \tiny{0.001} & $16.336 \pm $ \tiny{0.000} & $0.090 \pm $ \tiny{0.002} & $0.010 \pm $ \tiny{0.000} & $0.009 \pm $ \tiny{0.000} & $0.014 \pm $ \tiny{0.000} \\
 & PMLP~\cite{Yang2022Graph}              & $0.191 \pm $ \tiny{0.000} & $0.124 \pm $ \tiny{0.036} & $0.047 \pm $ \tiny{0.012} & $16.336 \pm $ \tiny{0.000} & $0.486 \pm $ \tiny{0.014} & $\bm{0.003 \pm }$ \textbf{\tiny{0.000}} & $0.008 \pm $ \tiny{0.001} & $0.013 \pm $ \tiny{0.000} \\
 & GraphSAGE~\cite{Hamilton2017Inductive} & $0.195 \pm $ \tiny{0.007} & $0.061 \pm $ \tiny{0.019} & $0.044 \pm $ \tiny{0.002} & $16.337 \pm $ \tiny{0.000} & $0.064 \pm $ \tiny{0.001} & $0.011 \pm $ \tiny{0.002} & $0.018 \pm $ \tiny{0.014} & $0.037 \pm $ \tiny{0.026} \\
 & GAT~\cite{velivckovic2018graph}         & $0.192 \pm $ \tiny{0.000} & $0.110 \pm $ \tiny{0.029} & $0.044 \pm $ \tiny{0.001} & $16.336 \pm $ \tiny{0.000} & $0.252 \pm $ \tiny{0.003} & $0.009 \pm $ \tiny{0.000} & $0.108 \pm $ \tiny{0.172} & $0.013 \pm $ \tiny{0.000} \\
 & GraphUNet~\cite{Gao2019Graph}          & $0.287 \pm $ \tiny{0.004} & $0.068 \pm $ \tiny{0.006} & $0.043 \pm $ \tiny{0.000} & $\bm{14.824 \pm }$ \textbf{\tiny{0.315}} & $0.085 \pm $ \tiny{0.002} & $0.009 \pm $ \tiny{0.000} & $0.009 \pm $ \tiny{0.000} & $0.013 \pm $ \tiny{0.000} \\
 & GIN~\cite{Xu2018How}                   & $0.336 \pm $ \tiny{0.005} & $0.069 \pm $ \tiny{0.040} & $0.043 \pm $ \tiny{0.000} & $16.336 \pm $ \tiny{0.000} & $0.491 \pm $ \tiny{0.038} & $0.009 \pm $ \tiny{0.000} & $0.009 \pm $ \tiny{0.000} & $0.013 \pm $ \tiny{0.000} \\
 & EdgeCNN~\cite{Wang2018Dynamic}         & $\bm{0.572 \pm }$ \textbf{\tiny{0.017}} & $0.072 \pm $ \tiny{0.047} & $\bm{0.158 \pm }$ \textbf{\tiny{0.035}} & $16.336 \pm $ \tiny{0.000} & $\bm{0.030 \pm }$ \textbf{\tiny{0.001}} & $0.007 \pm $ \tiny{0.009} & $\bm{0.006 \pm }$ \textbf{\tiny{0.000}} & $0.012 \pm $ \tiny{0.000} \\
    \bottomrule
    \end{tabular}
\end{table*}

{\bf Experimental set-up:}  
The hyperparameters for the GNN backbone models were chosen based on the reported values from each respective paper. When the  hyperparameters were not reported, we used the default values of the GCN~\cite{Kipf2016Semi}. Each experiment was repeated three times with different seeds, an early stopping with a patience of $50$ epochs was used, and a global training-time constraint of one hour per experiment was used to limit the computational costs. 

The complete \texttt{CHILI-3K} dataset was used in each experiment, whereas a comparable subset of $2975$ samples from the \texttt{CHILI-100K} dataset was used. The sub-sampling procedure was stratified based on \texttt{crystal\_system\_number}, resulting in each of the 7 classes being represented equally within the subset. The experimental setup for the individual models and tasks are described in more detail in Appendix \ref{app:modeltasksetup}. All the models were implemented in Pytorch~\cite{Paszke2019PyTorch}, and Pytorch-Geometric~\cite{Fey2019Fast} was used for the graph ML portions. 

\textbf{Property prediction tasks:} We propose three types of classification tasks for predicting: atom, crystal system and space group. The eight regression tasks for predicting: atom position, distance and six variations of scattering data from the nanomaterials (SAXS, SANS, XRD, ND, xPDF, nPDF). 

\textbf{Structure generation tasks:} To the best of our knowledge, no existing GNN models can be applied directly to the task of generating structures taking the available properties in the \texttt{CHILI}-datasets as input. We propose two simple formulations of the generative tasks, where the models receive one of the scattering data as input and is tasked with predicting (1) the fractional coordinates of the unit cell associated with that discrete particle and (2) the absolute atomic coordinates of the discrete particles. As a further simplification, note that, we ignore the prediction of any other node attributes.

\textbf{Metrics:} For evaluating the classification tasks, we use the weighted F1-score. For the regression tasks we use the mean absolute error (MAE) on the 3D positions for \texttt{pos\_abs} and mean squared error (MSE) for \texttt{edge\_attr}-, \texttt{saxs}-, \texttt{xrd}- and \texttt{xPDF}. 

\section{Results and discussion}

We next present the results for the different tasks described in Table~\ref{tab:tasks_overview}.
An overview of results on both {\tt CHILI} datasets for all the property prediction tasks are reported in Table \ref{tab:results_predictive},  and for the structure generation tasks in Table \ref{tab:results_generative}.  High-level trends observed for each task type are outlined next. Note that the methods in Table~\ref{tab:results_predictive} are sorted in increasing order of their complexity measured as the number of trainable parameters in the GNN-backbone (See Table~\ref{app:num_param} in Appendix~\ref{app:modeltasksetup} for exact numbers).

\textbf{Classification:} The naive predictors (random, MFC values) provide a useful baseline when comparing the more sophisticated GNN-based methods. In a majority of the scenarios, across the three classification tasks and the two datasets, all GNN-based methods do better than these naive baseline methods. In instances where the GNN methods do worse than these baselines, it could be explained by factors such as sub-optimal hyperparameters and limited training time.

For the \texttt{CHILI-3K} dataset, EdgeCNN~\cite{Wang2018Dynamic} which is the most complex model in terms of the number of parameters, consistently outperforms the other GNN-based methods across all the three classification tasks. This is an interesting trend which points to the usefulness of learning and aggregating features along the edges. This is in contrast to using node attributes along with the existing edge attributes as done with GCN~\cite{Kipf2016Semi} and GAT~\cite{velivckovic2018graph}. 

The \texttt{CHILI-100K} dataset which is more challenging compared to the {\tt CHILI-3K} dataset is reflected in the poor performance in all classification tasks by bulk of the GNN-based methods. In particular, \texttt{crystal\_system\_number} classification appears to be challenging for all the models.

\textbf{Regression:} Regression tasks are generally more difficult for deep learning models, due to factors such as incorrectly tuned bias parameters~\cite{Igel2023Remember}. Furthermore, the specific regression tasks formulated in these benchmarking tasks are more complex than the classification ones. This additional complexity could be due to the globalness of the prediction tasks, the dimensionality of the predictions, and lack of informative features in the input. The difficulty of these regression tasks is reflected across the board, including both {\tt CHILI} datasets, where the mean prediction seems to perform comparable to the GNN methods. This performance discrepancy between mean prediction and GNN-based methods is clearer for the {\tt pos\_abs} and {\tt edge\_attr} tasks.

For the regression of scattering data ({\tt saxs,xrd,xPDF}), the performance trends of the GNN-based models are slightly better. Methods like EdgeCNN~\cite{Wang2018Dynamic} again shows better performance compared to other methods. 
It is important to note that in the materials chemistry community the focus isn't typically on estimating scattering data from material graphs. This is primarily because the scattering data can easily be simulated, as demonstrated with the \texttt{CHILI}-datasets, using open-source software such as DebyeCalculator~\cite{Johansen2024GPU}. Consequently, the primary use of scattering data lies in the inverse task, such as using experimental scattering data to infer the atomic structure of the material. This being said, there are still potential value in accurately modelling scattering data from material graphs, particularly in the pursuit of discovering new materials. By doing so, researchers can explore the intricate relationship between structure and scattering data, facilitating data-driven approaches to material design. 
Examples of scattering data used in these prediction tasks can be found in Appendix~\ref{app:scattering_regression}.

\textbf{Structure generation:} The inverse materials design task is simplified as the prediction of {\tt pos\_abs} and \texttt{unit\_cell\_pos\_frac} features in the structure generation task. Reliable estimates of these two target variables can be used to generate structures.

For both \texttt{CHILI-3K} and \texttt{CHILI-100K}, the structure generation task show moderate errors across all different types of scattering data, as reported in Table~\ref{tab:results_generative}. The structure generation model generally performs better on the \texttt{CHILI-3K} dataset. The discrepancy between the two {\tt CHILI} datasets reflects the broader structural diversity present in the more complex \texttt{CHILI-100K} dataset. Sample reconstructions of predicted nanoparticles are provided in Appendix~\ref{app:generative}.

Structure generation with the \texttt{unit\_cell\_pos\_frac} target yields better reconstructions of the unit cell fractional coordinates. Notably on \texttt{CHILI-3K}, MAE values drop as low as $0.008$ using \texttt{xrd}. This outcome is expected, particularly on \texttt{CHILI-3K}, where the unit cell positions of each structure are tightly constrained, ranging from $0$ to $1$, and the unit cells themselves are mutually very distinct, which reduces the structure generation task to a form of classification into the 12 crystal systems visualized in Figure~\ref{fig:crystal_types}. Even then, a moderately higher MAE using \texttt{saxs} is to be expected. This is primarily because of the nature of SAXS data, which inherently lacks information on the unit cell parameters. The slightly higher MAE on \texttt{CHILI-100K} is due to the presence of moderately more irregular unit cells. Examples of unit cell reconstructions can be found in Appendix~\ref{app:generative}.

\begin{table}
    \caption{Overview of structure generation benchmark experiments on the \texttt{CHILI-3K} and \texttt{CHILI-100K} 
    datasets.}
    \vspace{-0.25cm}
    \label{tab:results_generative}
    \footnotesize
    \begin{tabular}{lcc}
        \toprule
        \multirow[c]{2}{*}{Task} & \texttt{CHILI-3K} & \texttt{CHILI-100K} \\
        \cmidrule{2-3}
         & \multicolumn{2}{c}{MAE ($\downarrow$)} \\
        \midrule
        \texttt{saxs} $\rightarrow$ \texttt{unit\_cell\_pos\_frac} & $0.053 \pm $ \tiny{0.009} & $0.198 \pm $ \tiny{0.003} \\
        \texttt{xrd} $\rightarrow$ \texttt{unit\_cell\_pos\_frac} & $0.008 \pm $ \tiny{0.002} & $0.191 \pm $ \tiny{0.000} \\
        \texttt{xPDF} $\rightarrow$ \texttt{unit\_cell\_pos\_frac} & $0.014 \pm $ \tiny{0.002} & $0.193 \pm $ \tiny{0.001}  \\
        \midrule
        \texttt{saxs} $\rightarrow$ \texttt{pos\_abs} & $1.539 \pm $ \tiny{0.007} \footnotesize{Å} & $2.783 \pm $ \tiny{0.002} \footnotesize{Å} \\
        \texttt{xrd} $\rightarrow$ \texttt{pos\_abs} & $1.894 \pm $ \tiny{0.014} \footnotesize{Å} & $2.779 \pm $ \tiny{0.006} \footnotesize{Å} \\
        \texttt{xPDF} $\rightarrow$ \texttt{pos\_abs} & $1.952 \pm $ \tiny{0.052} \footnotesize{Å} & $2.780 \pm $ \tiny{0.008} \footnotesize{Å} \\
        \bottomrule
    \end{tabular}
    \vspace{-0.35cm}
\end{table}

\section{Conclusion}

In this work we have presented a chemically-informed approach to generate large-scale graph datasets of nanomaterials. Using this approach we have provided two novel nanomaterial graph datasets: the medium-scale \texttt{CHILI-3K} dataset from mono-metal oxides with a wide range of interesting applications, and the large-scale \texttt{CHILI-100K} dataset from an experimental materials database with high structural diversity. 

The two \texttt{CHILI} datasets were benchmarked on 11 property prediction tasks using naive baselines and 7 off-the-shelf GNN models. The results show that all property prediction tasks are tractable, but also difficult enough that none of the off-the-shelf models are able to achieve usable performance for chemical applications. The EdgeCNN model performs best in general, which could be attributed to the layer architecture or that it has the most trainable parameters. Some tasks, like \texttt{atomic\_number} classification and \texttt{pos\_abs} regression, shows a tendency for the models getting stuck in local minima. We therefore think that future work focused on these tasks would be of high impact.

The datasets were also benchmarked on 6 structure prediction tasks using a simple property-to-structure model. The results show that, in a simplified setup, these tasks are tractable and harder to solve for discrete particles than the crystalline unit cells. This highlights the need for more work to address the challenges related to generation of variable size graphs and to achieve sub-Ångström positional precision, which would be useful for chemical applications.

{\bf Limitations:}
Both \texttt{CHILI} datasets assume that a discrete nanoparticle can be approximated using the unit cell from the crystalline material. This is not entirely physical, but we find it to be an acceptable trade-off for not having to relax all nanoparticles using DFT, which would require significant expertise and additional compute resources. 

Because of the way that \texttt{CHILI-3K} is constructed, the fractional atomic positions in the unit cells are identical across structures in each of crystal type subsets. 
Models trained for \texttt{unit\_cell\_pos\_frac} prediction tasks should therefore not be validated only on this dataset. We do not expect and have not observed any effects of this on the other tasks. 

The data in \texttt{CHILI-100K} is generated from a database of experimental materials, which has an inherent bias towards easily synthesized and stable crystalline materials. The data does, therefore, not cover all parts of the relevant chemical space equally and does not account for potential differences in stability between crystalline- and nanomaterials. This is especially important to consider if the \texttt{CHILI-100K} dataset is used for generating novel nanomaterials. 

The benchmark experiments were performed with limited hyperparameter tuning. The specific performance obtained by the GNN methods could be improved further by meticulous hyperparameter tuning or including additional tuning of the network architecture.

{\bf Open challenges and future work:} 
The prediction of 3D coordinates in {\em chemical systems} is an open problem in the chemistry and ML literature. This includes the chemical validity of the prediction and the consideration that the relative positions compared to the whole molecule or (nano)material are often more important than the absolute positions. We think the \texttt{CHILI} datasets can be an important contribution here with the multitude of complex positional data and the associated scattering data, which can be used as input signals, or measures for prediction quality.

Generative modelling of graphs with variable number of nodes and/or edges is an open problem in the graph ML literature. Solving this could revolutionize the field of generative graph ML, and thus also helping materials chemistry. One of the key limitations of the current state-of-the-art is the lack of datasets to facilitate scalability of graph generative models. We hope that the \texttt{CHILI} datasets can be an important contribution.

\begin{acks}
    This work is part of a project that has received funding from the European Research Council (ERC) under the European Union’s Horizon 2020 Research and Innovation Programme (grant agreement No. 804066). We are grateful for funding from University of Copenhagen through the Data+ program. 
    
    The authors also acknowledge participants of the Geilo Winter School on Graphs and Applications (Norway, 2024) for stress-testing an early subset of the data. 
\end{acks}

\bibliographystyle{ACM-Reference-Format}
\bibliography{references}


\begin{thebibliography}{91}


\ifx \showCODEN    \undefined \def \showCODEN     #1{\unskip}     \fi
\ifx \showDOI      \undefined \def \showDOI       #1{#1}\fi
\ifx \showISBNx    \undefined \def \showISBNx     #1{\unskip}     \fi
\ifx \showISBNxiii \undefined \def \showISBNxiii  #1{\unskip}     \fi
\ifx \showISSN     \undefined \def \showISSN      #1{\unskip}     \fi
\ifx \showLCCN     \undefined \def \showLCCN      #1{\unskip}     \fi
\ifx \shownote     \undefined \def \shownote      #1{#1}          \fi
\ifx \showarticletitle \undefined \def \showarticletitle #1{#1}   \fi
\ifx \showURL      \undefined \def \showURL       {\relax}        \fi
\providecommand\bibfield[2]{#2}
\providecommand\bibinfo[2]{#2}
\providecommand\natexlab[1]{#1}
\providecommand\showeprint[2][]{arXiv:#2}

\bibitem[Anker et~al\mbox{.}(2020)]%
        {Anker2020Characterising}
\bibfield{author}{\bibinfo{person}{Andy~S. Anker}, \bibinfo{person}{Emil T.~S.
  Kjær}, \bibinfo{person}{Erik~B. Dam}, \bibinfo{person}{Simon J.~L.
  Billinge}, \bibinfo{person}{Kirsten M.~Ø. Jensen}, {and}
  \bibinfo{person}{Raghavendra Selvan}.} \bibinfo{year}{2020}\natexlab{}.
\newblock \showarticletitle{Characterising the atomic structure of
  mono-metallic nanoparticles from x-ray scattering data using conditional
  generative models}. In \bibinfo{booktitle}{\emph{Proceedings of the 16th
  International Workshop on Mining and Learning with Graphs (MLG)}}.
\newblock


\bibitem[Banik et~al\mbox{.}(2023)]%
        {Banik2023CEGANN}
\bibfield{author}{\bibinfo{person}{Suvo Banik}, \bibinfo{person}{Debdas
  Dhabal}, \bibinfo{person}{Henry Chan}, \bibinfo{person}{Sukriti Manna},
  \bibinfo{person}{Mathew Cherukara}, \bibinfo{person}{Valeria Molinero}, {and}
  \bibinfo{person}{Subramanian K. R.~S. Sankaranarayanan}.}
  \bibinfo{year}{2023}\natexlab{}.
\newblock \showarticletitle{CEGANN: Crystal Edge Graph Attention Neural Network
  for multiscale classification of materials environment}.
\newblock \bibinfo{journal}{\emph{npj Computational Materials}}
  \bibinfo{volume}{9}, \bibinfo{number}{1} (\bibinfo{year}{2023}).
\newblock
\showISSN{2057-3960}


\bibitem[Barnard et~al\mbox{.}(2017)]%
        {Barnard2017Silver}
\bibfield{author}{\bibinfo{person}{Amanda Barnard}, \bibinfo{person}{Baichuan
  Sun}, \bibinfo{person}{Benyamin Motevalli~Soumehsaraei}, {and}
  \bibinfo{person}{George Opletal}.} \bibinfo{year}{2017}\natexlab{}.
\newblock \bibinfo{title}{Silver Nanoparticle Data Set}.
\newblock
\newblock


\bibitem[Batatia et~al\mbox{.}(2024)]%
        {Batatia2024Foundation}
\bibfield{author}{\bibinfo{person}{Ilyes Batatia}, \bibinfo{person}{Philipp
  Benner}, \bibinfo{person}{Yuan Chiang}, \bibinfo{person}{Alin~M. Elena},
  \bibinfo{person}{Dávid~P. Kovács}, \bibinfo{person}{Janosh Riebesell},
  \bibinfo{person}{Xavier~R. Advincula}, \bibinfo{person}{Mark Asta},
  \bibinfo{person}{William~J. Baldwin}, \bibinfo{person}{Noam Bernstein},
  \bibinfo{person}{Arghya Bhowmik}, \bibinfo{person}{Samuel~M. Blau},
  \bibinfo{person}{Vlad Cărare}, \bibinfo{person}{James~P. Darby},
  \bibinfo{person}{Sandip De}, \bibinfo{person}{Flaviano Della~Pia},
  \bibinfo{person}{Volker~L. Deringer}, \bibinfo{person}{Rokas Elijošius},
  \bibinfo{person}{Zakariya El-Machachi}, \bibinfo{person}{Edvin Fako},
  \bibinfo{person}{Andrea~C. Ferrari}, \bibinfo{person}{Annalena
  Genreith-Schriever}, \bibinfo{person}{Janine George}, \bibinfo{person}{Rhys
  E.~A. Goodall}, \bibinfo{person}{Clare~P. Grey}, \bibinfo{person}{Shuang
  Han}, \bibinfo{person}{Will Handley}, \bibinfo{person}{Hendrik~H. Heenen},
  \bibinfo{person}{Kersti Hermansson}, \bibinfo{person}{Christian Holm},
  \bibinfo{person}{Jad Jaafar}, \bibinfo{person}{Stephan Hofmann},
  \bibinfo{person}{Konstantin~S. Jakob}, \bibinfo{person}{Hyunwook Jung},
  \bibinfo{person}{Venkat Kapil}, \bibinfo{person}{Aaron~D. Kaplan},
  \bibinfo{person}{Nima Karimitari}, \bibinfo{person}{Namu Kroupa},
  \bibinfo{person}{Jolla Kullgren}, \bibinfo{person}{Matthew~C. Kuner},
  \bibinfo{person}{Domantas Kuryla}, \bibinfo{person}{Guoda Liepuoniute},
  \bibinfo{person}{Johannes~T. Margraf}, \bibinfo{person}{Ioan-Bogdan Magdău},
  \bibinfo{person}{Angelos Michaelides}, \bibinfo{person}{J.~Harry Moore},
  \bibinfo{person}{Aakash~A. Naik}, \bibinfo{person}{Samuel~P. Niblett},
  \bibinfo{person}{Sam~Walton Norwood}, \bibinfo{person}{Niamh O'Neill},
  \bibinfo{person}{Christoph Ortner}, \bibinfo{person}{Kristin~A. Persson},
  \bibinfo{person}{Karsten Reuter}, \bibinfo{person}{Andrew~S. Rosen},
  \bibinfo{person}{Lars~L. Schaaf}, \bibinfo{person}{Christoph Schran},
  \bibinfo{person}{Eric Sivonxay}, \bibinfo{person}{Tamás~K. Stenczel},
  \bibinfo{person}{Viktor Svahn}, \bibinfo{person}{Christopher Sutton},
  \bibinfo{person}{Cas van~der Oord}, \bibinfo{person}{Eszter Varga-Umbrich},
  \bibinfo{person}{Tejs Vegge}, \bibinfo{person}{Martin Vondrák},
  \bibinfo{person}{Yangshuai Wang}, \bibinfo{person}{William~C. Witt},
  \bibinfo{person}{Fabian Zills}, {and} \bibinfo{person}{Gábor Csányi}.}
  \bibinfo{year}{2024}\natexlab{}.
\newblock \showarticletitle{A foundation model for atomistic materials
  chemistry}.
\newblock  (\bibinfo{year}{2024}).
\newblock


\bibitem[Batatia et~al\mbox{.}(2022)]%
        {Batatia2022MACE}
\bibfield{author}{\bibinfo{person}{Ilyes Batatia}, \bibinfo{person}{David~P
  Kovacs}, \bibinfo{person}{Gregor Simm}, \bibinfo{person}{Christoph Ortner},
  {and} \bibinfo{person}{G{\'a}bor Cs{\'a}nyi}.}
  \bibinfo{year}{2022}\natexlab{}.
\newblock \showarticletitle{MACE: Higher order equivariant message passing
  neural networks for fast and accurate force fields}.
\newblock \bibinfo{journal}{\emph{Advances in Neural Information Processing
  Systems (NeurIPS)}} (\bibinfo{year}{2022}).
\newblock


\bibitem[Bouritsas et~al\mbox{.}(2023)]%
        {Bouritsas2023Improving}
\bibfield{author}{\bibinfo{person}{Giorgos Bouritsas},
  \bibinfo{person}{Fabrizio Frasca}, \bibinfo{person}{Stefanos Zafeiriou},
  {and} \bibinfo{person}{Michael~M. Bronstein}.}
  \bibinfo{year}{2023}\natexlab{}.
\newblock \showarticletitle{Improving Graph Neural Network Expressivity via
  Subgraph Isomorphism Counting}.
\newblock \bibinfo{journal}{\emph{IEEE Transactions on Pattern Analysis and
  Machine Intelligence}} \bibinfo{volume}{45}, \bibinfo{number}{1}
  (\bibinfo{year}{2023}), \bibinfo{pages}{657–668}.
\newblock
\showISSN{1939-3539}


\bibitem[Bronstein et~al\mbox{.}(2017)]%
        {Bronstein2017Geometric}
\bibfield{author}{\bibinfo{person}{Michael~M. Bronstein}, \bibinfo{person}{Joan
  Bruna}, \bibinfo{person}{Yann LeCun}, \bibinfo{person}{Arthur Szlam}, {and}
  \bibinfo{person}{Pierre Vandergheynst}.} \bibinfo{year}{2017}\natexlab{}.
\newblock \showarticletitle{Geometric Deep Learning: Going beyond Euclidean
  data}.
\newblock \bibinfo{journal}{\emph{IEEE Signal Processing Magazine}}
  \bibinfo{volume}{34}, \bibinfo{number}{4} (\bibinfo{year}{2017}),
  \bibinfo{pages}{18–42}.
\newblock
\showISSN{1558-0792}


\bibitem[Brown(1996)]%
        {Brown1996CIF}
\bibfield{author}{\bibinfo{person}{I.D. Brown}.}
  \bibinfo{year}{1996}\natexlab{}.
\newblock \showarticletitle{CIF (Crystallographic Information File). A standard
  for crystallographic data interchange}.
\newblock \bibinfo{journal}{\emph{Journal of Research of the National Institute
  of Standards and Technology}} \bibinfo{volume}{101}, \bibinfo{number}{3}
  (\bibinfo{year}{1996}), \bibinfo{pages}{341}.
\newblock
\showISSN{1044-677X}


\bibitem[Brus et~al\mbox{.}(1995)]%
        {Brus1995Electronic}
\bibfield{author}{\bibinfo{person}{L.~E. Brus}, \bibinfo{person}{P.~F.
  Szajowski}, \bibinfo{person}{W.~L. Wilson}, \bibinfo{person}{T.~D. Harris},
  \bibinfo{person}{S. Schuppler}, {and} \bibinfo{person}{P.~H. Citrin}.}
  \bibinfo{year}{1995}\natexlab{}.
\newblock \showarticletitle{Electronic Spectroscopy and Photophysics of Si
  Nanocrystals: Relationship to Bulk c-Si and Porous Si}.
\newblock \bibinfo{journal}{\emph{Journal of the American Chemical Society}}
  \bibinfo{volume}{117}, \bibinfo{number}{10} (\bibinfo{year}{1995}),
  \bibinfo{pages}{2915–2922}.
\newblock
\showISSN{1520-5126}


\bibitem[Cayley(1874)]%
        {Cayley1874LVII}
\bibfield{author}{\bibinfo{person}{Cayley}.} \bibinfo{year}{1874}\natexlab{}.
\newblock \showarticletitle{LVII. On the mathematical theory of isomers}.
\newblock \bibinfo{journal}{\emph{The London, Edinburgh, and Dublin
  Philosophical Magazine and Journal of Science}} \bibinfo{volume}{47},
  \bibinfo{number}{314} (\bibinfo{year}{1874}), \bibinfo{pages}{444–447}.
\newblock
\showISSN{1941-5990}


\bibitem[Chanussot et~al\mbox{.}(2021)]%
        {Chanussot2021Open}
\bibfield{author}{\bibinfo{person}{Lowik Chanussot}, \bibinfo{person}{Abhishek
  Das}, \bibinfo{person}{Siddharth Goyal}, \bibinfo{person}{Thibaut Lavril},
  \bibinfo{person}{Muhammed Shuaibi}, \bibinfo{person}{Morgane Riviere},
  \bibinfo{person}{Kevin Tran}, \bibinfo{person}{Javier Heras-Domingo},
  \bibinfo{person}{Caleb Ho}, \bibinfo{person}{Weihua Hu},
  \bibinfo{person}{Aini Palizhati}, \bibinfo{person}{Anuroop Sriram},
  \bibinfo{person}{Brandon Wood}, \bibinfo{person}{Junwoong Yoon},
  \bibinfo{person}{Devi Parikh}, \bibinfo{person}{C.~Lawrence Zitnick}, {and}
  \bibinfo{person}{Zachary Ulissi}.} \bibinfo{year}{2021}\natexlab{}.
\newblock \showarticletitle{Open Catalyst 2020 (OC20) Dataset and Community
  Challenges}.
\newblock \bibinfo{journal}{\emph{ACS Catalysis}} \bibinfo{volume}{11},
  \bibinfo{number}{10} (\bibinfo{year}{2021}), \bibinfo{pages}{6059–6072}.
\newblock
\showISSN{2155-5435}


\bibitem[Chen et~al\mbox{.}(2019)]%
        {Chen2019Graph}
\bibfield{author}{\bibinfo{person}{Chi Chen}, \bibinfo{person}{Weike Ye},
  \bibinfo{person}{Yunxing Zuo}, \bibinfo{person}{Chen Zheng}, {and}
  \bibinfo{person}{Shyue~Ping Ong}.} \bibinfo{year}{2019}\natexlab{}.
\newblock \showarticletitle{Graph Networks as a Universal Machine Learning
  Framework for Molecules and Crystals}.
\newblock \bibinfo{journal}{\emph{Chemistry of Materials}}
  \bibinfo{volume}{31}, \bibinfo{number}{9} (\bibinfo{year}{2019}),
  \bibinfo{pages}{3564–3572}.
\newblock
\showISSN{1520-5002}


\bibitem[Cheng et~al\mbox{.}(2021)]%
        {Cheng2021Geometric}
\bibfield{author}{\bibinfo{person}{Jiucheng Cheng}, \bibinfo{person}{Chunkai
  Zhang}, {and} \bibinfo{person}{Lifeng Dong}.}
  \bibinfo{year}{2021}\natexlab{}.
\newblock \showarticletitle{A geometric-information-enhanced crystal graph
  network for predicting properties of materials}.
\newblock \bibinfo{journal}{\emph{Communications Materials}}
  \bibinfo{volume}{2}, \bibinfo{number}{1} (\bibinfo{year}{2021}).
\newblock
\showISSN{2662-4443}


\bibitem[Choudhary and DeCost(2021)]%
        {Choudhary2021Atomistic}
\bibfield{author}{\bibinfo{person}{Kamal Choudhary} {and}
  \bibinfo{person}{Brian DeCost}.} \bibinfo{year}{2021}\natexlab{}.
\newblock \showarticletitle{Atomistic Line Graph Neural Network for improved
  materials property predictions}.
\newblock \bibinfo{journal}{\emph{npj Computational Materials}}
  \bibinfo{volume}{7}, \bibinfo{number}{1} (\bibinfo{year}{2021}).
\newblock
\showISSN{2057-3960}


\bibitem[Choudhary et~al\mbox{.}(2023)]%
        {Choudhary2023Large}
\bibfield{author}{\bibinfo{person}{Kamal Choudhary}, \bibinfo{person}{Daniel
  Wines}, \bibinfo{person}{Kangming Li}, \bibinfo{person}{Kevin~F. Garrity},
  \bibinfo{person}{Vishu Gupta}, \bibinfo{person}{Aldo~H. Romero},
  \bibinfo{person}{Jaron~T. Krogel}, \bibinfo{person}{Kayahan Saritas},
  \bibinfo{person}{Addis Fuhr}, \bibinfo{person}{Panchapakesan Ganesh},
  \bibinfo{person}{Paul R.~C. Kent}, \bibinfo{person}{Keqiang Yan},
  \bibinfo{person}{Yuchao Lin}, \bibinfo{person}{Shuiwang Ji},
  \bibinfo{person}{Ben Blaiszik}, \bibinfo{person}{Patrick Reiser},
  \bibinfo{person}{Pascal Friederich}, \bibinfo{person}{Ankit Agrawal},
  \bibinfo{person}{Pratyush Tiwary}, \bibinfo{person}{Eric Beyerle},
  \bibinfo{person}{Peter Minch}, \bibinfo{person}{Trevor~David Rhone},
  \bibinfo{person}{Ichiro Takeuchi}, \bibinfo{person}{Robert~B. Wexler},
  \bibinfo{person}{Arun Mannodi-Kanakkithodi}, \bibinfo{person}{Elif Ertekin},
  \bibinfo{person}{Avanish Mishra}, \bibinfo{person}{Nithin Mathew},
  \bibinfo{person}{Sterling~G. Baird}, \bibinfo{person}{Mitchell Wood},
  \bibinfo{person}{Andrew~Dale Rohskopf}, \bibinfo{person}{Jason
  Hattrick-Simpers}, \bibinfo{person}{Shih-Han Wang}, \bibinfo{person}{Luke
  E.~K. Achenie}, \bibinfo{person}{Hongliang Xin}, \bibinfo{person}{Maureen
  Williams}, \bibinfo{person}{Adam~J. Biacchi}, {and}
  \bibinfo{person}{Francesca Tavazza}.} \bibinfo{year}{2023}\natexlab{}.
\newblock \showarticletitle{Large Scale Benchmark of Materials Design Methods}.
\newblock  (\bibinfo{year}{2023}).
\newblock


\bibitem[Danish et~al\mbox{.}(2020)]%
        {Danish2020Systematic}
\bibfield{author}{\bibinfo{person}{Mir Sayed~Shah Danish},
  \bibinfo{person}{Arnab Bhattacharya}, \bibinfo{person}{Diana Stepanova},
  \bibinfo{person}{Alexey Mikhaylov}, \bibinfo{person}{Maria~Luisa Grilli},
  \bibinfo{person}{Mahdi Khosravy}, {and} \bibinfo{person}{Tomonobu Senjyu}.}
  \bibinfo{year}{2020}\natexlab{}.
\newblock \showarticletitle{A Systematic Review of Metal Oxide Applications for
  Energy and Environmental Sustainability}.
\newblock \bibinfo{journal}{\emph{Metals}} \bibinfo{volume}{10},
  \bibinfo{number}{12} (\bibinfo{year}{2020}), \bibinfo{pages}{1604}.
\newblock
\showISSN{2075-4701}


\bibitem[Das et~al\mbox{.}(2023)]%
        {Das2023CrysGNN}
\bibfield{author}{\bibinfo{person}{Kishalay Das}, \bibinfo{person}{Bidisha
  Samanta}, \bibinfo{person}{Pawan Goyal}, \bibinfo{person}{Seung-Cheol Lee},
  \bibinfo{person}{Satadeep Bhattacharjee}, {and} \bibinfo{person}{Niloy
  Ganguly}.} \bibinfo{year}{2023}\natexlab{}.
\newblock \showarticletitle{Crys{GNN} : Distilling pre-trained knowledge to
  enhance property prediction for crystalline materials.}. In
  \bibinfo{booktitle}{\emph{Workshop on ''Machine Learning for Materials'' at
  ICLR 2023}}.
\newblock


\bibitem[De~Cao and Kipf(2018)]%
        {DeCao2018MolGAN}
\bibfield{author}{\bibinfo{person}{Nicola De~Cao} {and} \bibinfo{person}{Thomas
  Kipf}.} \bibinfo{year}{2018}\natexlab{}.
\newblock \showarticletitle{{MolGAN: An implicit generative model for small
  molecular graphs}}.
\newblock \bibinfo{journal}{\emph{ICML 2018 workshop on Theoretical Foundations
  and Applications of Deep Generative Models}} (\bibinfo{year}{2018}).
\newblock


\bibitem[Downs and Hall-Wallace(2003)]%
        {Downs2003TheAM}
\bibfield{author}{\bibinfo{person}{Robert~T. Downs} {and}
  \bibinfo{person}{Michelle Hall-Wallace}.} \bibinfo{year}{2003}\natexlab{}.
\newblock \showarticletitle{The American Mineralogist crystal structure
  database}.
\newblock \bibinfo{journal}{\emph{American Mineralogist}}  \bibinfo{volume}{88}
  (\bibinfo{year}{2003}), \bibinfo{pages}{247--250}.
\newblock


\bibitem[Duvenaud et~al\mbox{.}(2015)]%
        {Duvenaud2015Convolutional}
\bibfield{author}{\bibinfo{person}{David~K Duvenaud}, \bibinfo{person}{Dougal
  Maclaurin}, \bibinfo{person}{Jorge Iparraguirre}, \bibinfo{person}{Rafael
  Bombarell}, \bibinfo{person}{Timothy Hirzel}, \bibinfo{person}{Alan
  Aspuru-Guzik}, {and} \bibinfo{person}{Ryan~P Adams}.}
  \bibinfo{year}{2015}\natexlab{}.
\newblock \showarticletitle{Convolutional Networks on Graphs for Learning
  Molecular Fingerprints}. In \bibinfo{booktitle}{\emph{Advances in Neural
  Information Processing Systems}}, Vol.~\bibinfo{volume}{28}.
\newblock


\bibitem[Fey and Lenssen(2019)]%
        {Fey2019Fast}
\bibfield{author}{\bibinfo{person}{Matthias Fey} {and}
  \bibinfo{person}{Jan~Eric Lenssen}.} \bibinfo{year}{2019}\natexlab{}.
\newblock \bibinfo{journal}{\emph{ICML 2018 workshop on Representation Learning
  on Graphs and Manifolds}} (\bibinfo{year}{2019}).
\newblock


\bibitem[Fung and Jiang(2017)]%
        {Fung2017Exploring}
\bibfield{author}{\bibinfo{person}{Victor Fung} {and} \bibinfo{person}{De-en
  Jiang}.} \bibinfo{year}{2017}\natexlab{}.
\newblock \showarticletitle{Exploring Structural Diversity and Fluxionality of
  Ptn (n = 10–13) Clusters from First-Principles}.
\newblock \bibinfo{journal}{\emph{The Journal of Physical Chemistry C}}
  \bibinfo{volume}{121}, \bibinfo{number}{20} (\bibinfo{year}{2017}),
  \bibinfo{pages}{10796–10802}.
\newblock
\showISSN{1932-7455}


\bibitem[Fung et~al\mbox{.}(2021)]%
        {fung2021benchmarking}
\bibfield{author}{\bibinfo{person}{Victor Fung}, \bibinfo{person}{Jiaxin
  Zhang}, \bibinfo{person}{Eric Juarez}, {and} \bibinfo{person}{Bobby~G.
  Sumpter}.} \bibinfo{year}{2021}\natexlab{}.
\newblock \showarticletitle{Benchmarking graph neural networks for materials
  chemistry}.
\newblock \bibinfo{journal}{\emph{npj Computational Materials}}
  \bibinfo{volume}{7}, \bibinfo{number}{1} (\bibinfo{year}{2021}).
\newblock
\showISSN{2057-3960}


\bibitem[Ganachari et~al\mbox{.}(2019)]%
        {Ganachari2019Metal}
\bibfield{author}{\bibinfo{person}{Sharanabasava~V. Ganachari},
  \bibinfo{person}{Leena Hublikar}, \bibinfo{person}{Jayachandra~S. Yaradoddi},
  {and} \bibinfo{person}{Shivalingayya~S. Math}.}
  \bibinfo{year}{2019}\natexlab{}.
\newblock \bibinfo{booktitle}{\emph{Metal Oxide Nanomaterials for Environmental
  Applications}}.
\newblock \bibinfo{pages}{2357–2368}.
\newblock


\bibitem[Gao and Ji(2019)]%
        {Gao2019Graph}
\bibfield{author}{\bibinfo{person}{Hongyang Gao} {and}
  \bibinfo{person}{Shuiwang Ji}.} \bibinfo{year}{2019}\natexlab{}.
\newblock \showarticletitle{{Graph U-nets}}. In
  \bibinfo{booktitle}{\emph{International Conference on Machine Learning
  (ICML)}}.
\newblock


\bibitem[Gilmer et~al\mbox{.}(2017)]%
        {Gilmer2017Neural}
\bibfield{author}{\bibinfo{person}{Justin Gilmer}, \bibinfo{person}{Samuel~S.
  Schoenholz}, \bibinfo{person}{Patrick~F. Riley}, \bibinfo{person}{Oriol
  Vinyals}, {and} \bibinfo{person}{George~E. Dahl}.}
  \bibinfo{year}{2017}\natexlab{}.
\newblock \showarticletitle{Neural Message Passing for Quantum Chemistry}. In
  \bibinfo{booktitle}{\emph{Proceedings of the 34th International Conference on
  Machine Learning}}, Vol.~\bibinfo{volume}{70}. \bibinfo{pages}{1263--1272}.
\newblock


\bibitem[Gong et~al\mbox{.}(2023)]%
        {Gong2022Examining}
\bibfield{author}{\bibinfo{person}{Sheng Gong}, \bibinfo{person}{Keqiang Yan},
  \bibinfo{person}{Tian Xie}, \bibinfo{person}{Yang Shao-Horn},
  \bibinfo{person}{Rafael Gomez-Bombarelli}, \bibinfo{person}{Shuiwang Ji},
  {and} \bibinfo{person}{Jeffrey~C Grossman}.} \bibinfo{year}{2023}\natexlab{}.
\newblock \showarticletitle{Examining graph neural networks for crystal
  structures: limitations and opportunities for capturing periodicity}.
\newblock \bibinfo{journal}{\emph{Science Advances}} (\bibinfo{year}{2023}).
\newblock


\bibitem[Goodall and Lee(2020)]%
        {Goodall2020Predicting}
\bibfield{author}{\bibinfo{person}{Rhys E.~A. Goodall} {and}
  \bibinfo{person}{Alpha~A. Lee}.} \bibinfo{year}{2020}\natexlab{}.
\newblock \showarticletitle{Predicting materials properties without crystal
  structure: deep representation learning from stoichiometry}.
\newblock \bibinfo{journal}{\emph{Nature Communications}} \bibinfo{volume}{11},
  \bibinfo{number}{1} (\bibinfo{year}{2020}).
\newblock
\showISSN{2041-1723}


\bibitem[Gražulis et~al\mbox{.}(2009)]%
        {Graulis2009Crystallography}
\bibfield{author}{\bibinfo{person}{Saulius Gražulis}, \bibinfo{person}{Daniel
  Chateigner}, \bibinfo{person}{Robert~T. Downs}, \bibinfo{person}{A.~F.~T.
  Yokochi}, \bibinfo{person}{Miguel Quirós}, \bibinfo{person}{Luca
  Lutterotti}, \bibinfo{person}{Elena Manakova}, \bibinfo{person}{Justas
  Butkus}, \bibinfo{person}{Peter Moeck}, {and} \bibinfo{person}{Armel
  Le~Bail}.} \bibinfo{year}{2009}\natexlab{}.
\newblock \showarticletitle{Crystallography Open Database – an open-access
  collection of crystal structures}.
\newblock \bibinfo{journal}{\emph{Journal of Applied Crystallography}}
  \bibinfo{volume}{42}, \bibinfo{number}{4} (\bibinfo{year}{2009}),
  \bibinfo{pages}{726–729}.
\newblock
\showISSN{0021-8898}


\bibitem[Gražulis et~al\mbox{.}(2011)]%
        {Graulis2011Crystallography}
\bibfield{author}{\bibinfo{person}{Saulius Gražulis}, \bibinfo{person}{Adriana
  Daškevič}, \bibinfo{person}{Andrius Merkys}, \bibinfo{person}{Daniel
  Chateigner}, \bibinfo{person}{Luca Lutterotti}, \bibinfo{person}{Miguel
  Quirós}, \bibinfo{person}{Nadezhda~R. Serebryanaya}, \bibinfo{person}{Peter
  Moeck}, \bibinfo{person}{Robert~T. Downs}, {and} \bibinfo{person}{Armel
  Le~Bail}.} \bibinfo{year}{2011}\natexlab{}.
\newblock \showarticletitle{Crystallography Open Database (COD): an open-access
  collection of crystal structures and platform for world-wide collaboration}.
\newblock \bibinfo{journal}{\emph{Nucleic Acids Research}}
  \bibinfo{volume}{40}, \bibinfo{number}{D1} (\bibinfo{year}{2011}),
  \bibinfo{pages}{D420–D427}.
\newblock
\showISSN{0305-1048}


\bibitem[Gražulis et~al\mbox{.}(2015)]%
        {Graulis2015Computing}
\bibfield{author}{\bibinfo{person}{Saulius Gražulis}, \bibinfo{person}{Andrius
  Merkys}, \bibinfo{person}{Antanas Vaitkus}, {and} \bibinfo{person}{Mykolas
  Okulič-Kazarinas}.} \bibinfo{year}{2015}\natexlab{}.
\newblock \showarticletitle{Computing stoichiometric molecular composition from
  crystal structures}.
\newblock \bibinfo{journal}{\emph{Journal of Applied Crystallography}}
  \bibinfo{volume}{48}, \bibinfo{number}{1} (\bibinfo{year}{2015}),
  \bibinfo{pages}{85–91}.
\newblock
\showISSN{1600-5767}


\bibitem[Groom et~al\mbox{.}(2016)]%
        {Groom2016Cambridge}
\bibfield{author}{\bibinfo{person}{Colin~R. Groom}, \bibinfo{person}{Ian~J.
  Bruno}, \bibinfo{person}{Matthew~P. Lightfoot}, {and}
  \bibinfo{person}{Suzanna~C. Ward}.} \bibinfo{year}{2016}\natexlab{}.
\newblock \showarticletitle{The Cambridge Structural Database}.
\newblock \bibinfo{journal}{\emph{Acta Crystallographica Section B Structural
  Science, Crystal Engineering and Materials}} \bibinfo{volume}{72},
  \bibinfo{number}{2} (\bibinfo{year}{2016}), \bibinfo{pages}{171–179}.
\newblock
\showISSN{2052-5206}


\bibitem[Hamilton et~al\mbox{.}(2017)]%
        {Hamilton2017Inductive}
\bibfield{author}{\bibinfo{person}{Will Hamilton}, \bibinfo{person}{Zhitao
  Ying}, {and} \bibinfo{person}{Jure Leskovec}.}
  \bibinfo{year}{2017}\natexlab{}.
\newblock \showarticletitle{Inductive representation learning on large graphs}.
\newblock \bibinfo{journal}{\emph{Advances in Neural Information Processing
  Systems (NeurIPS)}} (\bibinfo{year}{2017}).
\newblock


\bibitem[Hamilton(2020)]%
        {Hamilton2020Graph}
\bibfield{author}{\bibinfo{person}{William~L. Hamilton}.}
  \bibinfo{year}{2020}\natexlab{}.
\newblock \showarticletitle{Graph Representation Learning}.
\newblock \bibinfo{journal}{\emph{Synthesis Lectures on Artificial Intelligence
  and Machine Learning}} \bibinfo{volume}{14}, \bibinfo{number}{3}
  (\bibinfo{year}{2020}), \bibinfo{pages}{1--159}.
\newblock


\bibitem[Hjorth~Larsen et~al\mbox{.}(2017)]%
        {HjorthLarsen2017Atomic}
\bibfield{author}{\bibinfo{person}{Ask Hjorth~Larsen}, \bibinfo{person}{Jens
  Jørgen~Mortensen}, \bibinfo{person}{Jakob Blomqvist},
  \bibinfo{person}{Ivano~E Castelli}, \bibinfo{person}{Rune Christensen},
  \bibinfo{person}{Marcin Dułak}, \bibinfo{person}{Jesper Friis},
  \bibinfo{person}{Michael~N Groves}, \bibinfo{person}{Bjørk Hammer},
  \bibinfo{person}{Cory Hargus}, \bibinfo{person}{Eric~D Hermes},
  \bibinfo{person}{Paul~C Jennings}, \bibinfo{person}{Peter Bjerre~Jensen},
  \bibinfo{person}{James Kermode}, \bibinfo{person}{John~R Kitchin},
  \bibinfo{person}{Esben Leonhard~Kolsbjerg}, \bibinfo{person}{Joseph Kubal},
  \bibinfo{person}{Kristen Kaasbjerg}, \bibinfo{person}{Steen Lysgaard},
  \bibinfo{person}{Jón Bergmann~Maronsson}, \bibinfo{person}{Tristan Maxson},
  \bibinfo{person}{Thomas Olsen}, \bibinfo{person}{Lars Pastewka},
  \bibinfo{person}{Andrew Peterson}, \bibinfo{person}{Carsten Rostgaard},
  \bibinfo{person}{Jakob Schiøtz}, \bibinfo{person}{Ole Schütt},
  \bibinfo{person}{Mikkel Strange}, \bibinfo{person}{Kristian~S Thygesen},
  \bibinfo{person}{Tejs Vegge}, \bibinfo{person}{Lasse Vilhelmsen},
  \bibinfo{person}{Michael Walter}, \bibinfo{person}{Zhenhua Zeng}, {and}
  \bibinfo{person}{Karsten~W Jacobsen}.} \bibinfo{year}{2017}\natexlab{}.
\newblock \showarticletitle{The atomic simulation environment—a Python
  library for working with atoms}.
\newblock \bibinfo{journal}{\emph{Journal of Physics: Condensed Matter}}
  \bibinfo{volume}{29}, \bibinfo{number}{27} (\bibinfo{year}{2017}),
  \bibinfo{pages}{273002}.
\newblock
\showISSN{1361-648X}


\bibitem[Ho et~al\mbox{.}(2020)]%
        {Ho2020Denoising}
\bibfield{author}{\bibinfo{person}{Jonathan Ho}, \bibinfo{person}{Ajay Jain},
  {and} \bibinfo{person}{Pieter Abbeel}.} \bibinfo{year}{2020}\natexlab{}.
\newblock \showarticletitle{Denoising Diffusion Probabilistic Models}. In
  \bibinfo{booktitle}{\emph{Advances in Neural Information Processing
  Systems}}, Vol.~\bibinfo{volume}{33}. \bibinfo{pages}{6840--6851}.
\newblock


\bibitem[Hu et~al\mbox{.}(2020)]%
        {Hu2020Open}
\bibfield{author}{\bibinfo{person}{Weihua Hu}, \bibinfo{person}{Matthias Fey},
  \bibinfo{person}{Marinka Zitnik}, \bibinfo{person}{Yuxiao Dong},
  \bibinfo{person}{Hongyu Ren}, \bibinfo{person}{Bowen Liu},
  \bibinfo{person}{Michele Catasta}, {and} \bibinfo{person}{Jure Leskovec}.}
  \bibinfo{year}{2020}\natexlab{}.
\newblock \showarticletitle{Open Graph Benchmark: Datasets for Machine Learning
  on Graphs}. In \bibinfo{booktitle}{\emph{Advances in Neural Information
  Processing Systems}}, Vol.~\bibinfo{volume}{33}.
  \bibinfo{pages}{22118--22133}.
\newblock


\bibitem[Hussain et~al\mbox{.}(2022)]%
        {Hussain2022Global}
\bibfield{author}{\bibinfo{person}{Md~Shamim Hussain},
  \bibinfo{person}{Mohammed~J. Zaki}, {and} \bibinfo{person}{Dharmashankar
  Subramanian}.} \bibinfo{year}{2022}\natexlab{}.
\newblock \showarticletitle{Global Self-Attention as a Replacement for Graph
  Convolution}. In \bibinfo{booktitle}{\emph{Proceedings of the 28th ACM SIGKDD
  Conference on Knowledge Discovery and Data Mining}}.
\newblock


\bibitem[Igel and Oehmcke(2023)]%
        {Igel2023Remember}
\bibfield{author}{\bibinfo{person}{Christian Igel} {and}
  \bibinfo{person}{Stefan Oehmcke}.} \bibinfo{year}{2023}\natexlab{}.
\newblock \showarticletitle{Remember to Correct the Bias When Using Deep
  Learning for Regression!}
\newblock \bibinfo{journal}{\emph{KI - K\"{u}nstliche Intelligenz}}
  \bibinfo{volume}{37}, \bibinfo{number}{1} (\bibinfo{year}{2023}),
  \bibinfo{pages}{33–40}.
\newblock
\showISSN{1610-1987}


\bibitem[Jain et~al\mbox{.}(2013)]%
        {Jain2013Commentary}
\bibfield{author}{\bibinfo{person}{Anubhav Jain}, \bibinfo{person}{Shyue~Ping
  Ong}, \bibinfo{person}{Geoffroy Hautier}, \bibinfo{person}{Wei Chen},
  \bibinfo{person}{William~Davidson Richards}, \bibinfo{person}{Stephen Dacek},
  \bibinfo{person}{Shreyas Cholia}, \bibinfo{person}{Dan Gunter},
  \bibinfo{person}{David Skinner}, \bibinfo{person}{Gerbrand Ceder}, {and}
  \bibinfo{person}{Kristin~A. Persson}.} \bibinfo{year}{2013}\natexlab{}.
\newblock \showarticletitle{Commentary: The Materials Project: A materials
  genome approach to accelerating materials innovation}.
\newblock \bibinfo{journal}{\emph{APL Materials}} \bibinfo{volume}{1},
  \bibinfo{number}{1} (\bibinfo{year}{2013}).
\newblock
\showISSN{2166-532X}


\bibitem[Jha et~al\mbox{.}(2018)]%
        {Jha2018ElemNet}
\bibfield{author}{\bibinfo{person}{Dipendra Jha}, \bibinfo{person}{Logan Ward},
  \bibinfo{person}{Arindam Paul}, \bibinfo{person}{Wei-keng Liao},
  \bibinfo{person}{Alok Choudhary}, \bibinfo{person}{Chris Wolverton}, {and}
  \bibinfo{person}{Ankit Agrawal}.} \bibinfo{year}{2018}\natexlab{}.
\newblock \showarticletitle{ElemNet: Deep Learning the Chemistry of Materials
  From Only Elemental Composition}.
\newblock \bibinfo{journal}{\emph{Scientific Reports}} \bibinfo{volume}{8},
  \bibinfo{number}{1} (\bibinfo{year}{2018}).
\newblock
\showISSN{2045-2322}


\bibitem[Jha et~al\mbox{.}(2019)]%
        {Jha2019IRNet}
\bibfield{author}{\bibinfo{person}{Dipendra Jha}, \bibinfo{person}{Logan Ward},
  \bibinfo{person}{Zijiang Yang}, \bibinfo{person}{Christopher Wolverton},
  \bibinfo{person}{Ian Foster}, \bibinfo{person}{Wei-keng Liao},
  \bibinfo{person}{Alok Choudhary}, {and} \bibinfo{person}{Ankit Agrawal}.}
  \bibinfo{year}{2019}\natexlab{}.
\newblock \showarticletitle{IRNet: A General Purpose Deep Residual Regression
  Framework for Materials Discovery}. In \bibinfo{booktitle}{\emph{Proceedings
  of the 25th ACM SIGKDD International Conference on Knowledge Discovery \&
  Data Mining}}.
\newblock


\bibitem[Johansen et~al\mbox{.}(2024)]%
        {Johansen2024GPU}
\bibfield{author}{\bibinfo{person}{Frederik~L. Johansen},
  \bibinfo{person}{Andy~S. Anker}, \bibinfo{person}{Ulrik Friis-Jensen},
  \bibinfo{person}{Erik~B. Dam}, \bibinfo{person}{Kirsten M.~Ø. Jensen}, {and}
  \bibinfo{person}{Raghavendra Selvan}.} \bibinfo{year}{2024}\natexlab{}.
\newblock \showarticletitle{{A GPU-Accelerated Open-Source Python Package for
  Calculating Powder Diffraction, Small-Angle-, and Total Scattering with the
  Debye Scattering Equation}}.
\newblock \bibinfo{journal}{\emph{Journal of Open Source Software}}
  (\bibinfo{year}{2024}).
\newblock


\bibitem[Jung et~al\mbox{.}(2023)]%
        {Jung2023High}
\bibfield{author}{\bibinfo{person}{Jong~Hyun Jung}, \bibinfo{person}{Prashanth
  Srinivasan}, \bibinfo{person}{Axel Forslund}, {and} \bibinfo{person}{Blazej
  Grabowski}.} \bibinfo{year}{2023}\natexlab{}.
\newblock \showarticletitle{High-accuracy thermodynamic properties to the
  melting point from ab initio calculations aided by machine-learning
  potentials}.
\newblock \bibinfo{journal}{\emph{npj Computational Materials}}
  \bibinfo{volume}{9}, \bibinfo{number}{1} (\bibinfo{year}{2023}).
\newblock
\showISSN{2057-3960}


\bibitem[Jurs(1971)]%
        {Jurs1971Machine}
\bibfield{author}{\bibinfo{person}{Peter~C. Jurs}.}
  \bibinfo{year}{1971}\natexlab{}.
\newblock \showarticletitle{Machine Intelligence Applied to Chemical Systems: A
  Graph Theoretical and Learning Machine Study of Second-Order Effects in Low
  Resolution Mass Spectra}.
\newblock \bibinfo{journal}{\emph{Applied Spectroscopy}} \bibinfo{volume}{25},
  \bibinfo{number}{4} (\bibinfo{year}{1971}), \bibinfo{pages}{483–488}.
\newblock
\showISSN{1943-3530}


\bibitem[Kingma and Ba(2015)]%
        {Kingma2014Adam}
\bibfield{author}{\bibinfo{person}{Diederik~P. Kingma} {and}
  \bibinfo{person}{Jimmy Ba}.} \bibinfo{year}{2015}\natexlab{}.
\newblock \showarticletitle{Adam: A Method for Stochastic Optimization}. In
  \bibinfo{booktitle}{\emph{International Conference on Learning
  Representations (ICLR)}}.
\newblock


\bibitem[Kipf and Welling(2017)]%
        {Kipf2016Semi}
\bibfield{author}{\bibinfo{person}{Thomas~N Kipf} {and} \bibinfo{person}{Max
  Welling}.} \bibinfo{year}{2017}\natexlab{}.
\newblock \showarticletitle{Semi-Supervised Classification with Graph
  Convolutional Networks}. In \bibinfo{booktitle}{\emph{International
  Conference on Learning Representations (ICLR)}}.
\newblock


\bibitem[Kirklin et~al\mbox{.}(2015)]%
        {Kirklin2015Open}
\bibfield{author}{\bibinfo{person}{Scott Kirklin}, \bibinfo{person}{James~E
  Saal}, \bibinfo{person}{Bryce Meredig}, \bibinfo{person}{Alex Thompson},
  \bibinfo{person}{Jeff~W Doak}, \bibinfo{person}{Muratahan Aykol},
  \bibinfo{person}{Stephan R\"{u}hl}, {and} \bibinfo{person}{Chris Wolverton}.}
  \bibinfo{year}{2015}\natexlab{}.
\newblock \showarticletitle{The Open Quantum Materials Database (OQMD):
  assessing the accuracy of DFT formation energies}.
\newblock \bibinfo{journal}{\emph{npj Computational Materials}}
  \bibinfo{volume}{1}, \bibinfo{number}{1} (\bibinfo{year}{2015}).
\newblock
\showISSN{2057-3960}


\bibitem[Kjær et~al\mbox{.}(2023)]%
        {Kjaer2023DeepStruc}
\bibfield{author}{\bibinfo{person}{Emil T.~S. Kjær}, \bibinfo{person}{Andy~S.
  Anker}, \bibinfo{person}{Marcus~N. Weng}, \bibinfo{person}{Simon J.~L.
  Billinge}, \bibinfo{person}{Raghavendra Selvan}, {and}
  \bibinfo{person}{Kirsten M.~Ø. Jensen}.} \bibinfo{year}{2023}\natexlab{}.
\newblock \showarticletitle{DeepStruc: towards structure solution from pair
  distribution function data using deep generative models}.
\newblock \bibinfo{journal}{\emph{Digital Discovery}} \bibinfo{volume}{2},
  \bibinfo{number}{1} (\bibinfo{year}{2023}), \bibinfo{pages}{69–80}.
\newblock
\showISSN{2635-098X}


\bibitem[Laurent et~al\mbox{.}(2018)]%
        {Laurent2018Metal}
\bibfield{author}{\bibinfo{person}{S. Laurent}, \bibinfo{person}{S. Boutry},
  {and} \bibinfo{person}{R.N. Muller}.} \bibinfo{year}{2018}\natexlab{}.
\newblock \bibinfo{booktitle}{\emph{Metal Oxide Particles and Their Prospects
  for Applications}}.
\newblock \bibinfo{pages}{3–42}.
\newblock
\showISBNx{9780081019252}


\bibitem[LeCun et~al\mbox{.}(2015)]%
        {LeCun2015Deep}
\bibfield{author}{\bibinfo{person}{Yann LeCun}, \bibinfo{person}{Yoshua
  Bengio}, {and} \bibinfo{person}{Geoffrey Hinton}.}
  \bibinfo{year}{2015}\natexlab{}.
\newblock \showarticletitle{Deep learning}.
\newblock \bibinfo{journal}{\emph{Nature}} \bibinfo{volume}{521},
  \bibinfo{number}{7553} (\bibinfo{year}{2015}), \bibinfo{pages}{436–444}.
\newblock
\showISSN{1476-4687}


\bibitem[Lee et~al\mbox{.}(2023)]%
        {Lee2023MatSciML}
\bibfield{author}{\bibinfo{person}{Kin Long~Kelvin Lee},
  \bibinfo{person}{Carmelo Gonzales}, \bibinfo{person}{Marcel Nassar},
  \bibinfo{person}{Matthew Spellings}, \bibinfo{person}{Mikhail Galkin}, {and}
  \bibinfo{person}{Santiago Miret}.} \bibinfo{year}{2023}\natexlab{}.
\newblock \showarticletitle{MatSci{ML}: A Broad, Multi-Task Benchmark for
  Solid-State Materials Modeling}. In \bibinfo{booktitle}{\emph{AI for
  Accelerated Materials Design - NeurIPS 2023 Workshop}}.
\newblock


\bibitem[Leskovec and Mcauley(2012)]%
        {Leskovec2012Learning}
\bibfield{author}{\bibinfo{person}{Jure Leskovec} {and} \bibinfo{person}{Julian
  Mcauley}.} \bibinfo{year}{2012}\natexlab{}.
\newblock \showarticletitle{Learning to Discover Social Circles in Ego
  Networks}. In \bibinfo{booktitle}{\emph{Advances in Neural Information
  Processing Systems}}, Vol.~\bibinfo{volume}{25}. \bibinfo{publisher}{Curran
  Associates, Inc.}
\newblock


\bibitem[Liu et~al\mbox{.}(2021)]%
        {Liu2021GraphEBM}
\bibfield{author}{\bibinfo{person}{Meng Liu}, \bibinfo{person}{Keqiang Yan},
  \bibinfo{person}{Bora Oztekin}, {and} \bibinfo{person}{Shuiwang Ji}.}
  \bibinfo{year}{2021}\natexlab{}.
\newblock \showarticletitle{Graph{EBM}: Molecular Graph Generation with
  Energy-Based Models}. In \bibinfo{booktitle}{\emph{Energy Based Models
  Workshop - ICLR 2021}}.
\newblock


\bibitem[Louis et~al\mbox{.}(2020)]%
        {Louis2020Graph}
\bibfield{author}{\bibinfo{person}{Steph-Yves Louis}, \bibinfo{person}{Yong
  Zhao}, \bibinfo{person}{Alireza Nasiri}, \bibinfo{person}{Xiran Wang},
  \bibinfo{person}{Yuqi Song}, \bibinfo{person}{Fei Liu}, {and}
  \bibinfo{person}{Jianjun Hu}.} \bibinfo{year}{2020}\natexlab{}.
\newblock \showarticletitle{Graph convolutional neural networks with global
  attention for improved materials property prediction}.
\newblock \bibinfo{journal}{\emph{Physical Chemistry Chemical Physics}}
  \bibinfo{volume}{22}, \bibinfo{number}{32} (\bibinfo{year}{2020}),
  \bibinfo{pages}{18141–18148}.
\newblock
\showISSN{1463-9084}


\bibitem[Manna et~al\mbox{.}(2023)]%
        {Manna2023Database}
\bibfield{author}{\bibinfo{person}{Sukriti Manna}, \bibinfo{person}{Yunzhe
  Wang}, \bibinfo{person}{Alberto Hernandez}, \bibinfo{person}{Peter Lile},
  \bibinfo{person}{Shanping Liu}, {and} \bibinfo{person}{Tim Mueller}.}
  \bibinfo{year}{2023}\natexlab{}.
\newblock \showarticletitle{A database of low-energy atomically precise
  nanoclusters}.
\newblock \bibinfo{journal}{\emph{Scientific Data}} \bibinfo{volume}{10},
  \bibinfo{number}{1} (\bibinfo{year}{2023}).
\newblock
\showISSN{2052-4463}


\bibitem[Mentel(2023)]%
        {Mentel2023Mendeleev}
\bibfield{author}{\bibinfo{person}{Łukasz Mentel}.}
  \bibinfo{year}{2023}\natexlab{}.
\newblock \bibinfo{booktitle}{\emph{{mendeleev - A Python package with
  properties of chemical elements, ions, isotopes and methods to manipulate and
  visualize periodic table.}}}
\newblock


\bibitem[Merchant et~al\mbox{.}(2023)]%
        {Merchant2023Scaling}
\bibfield{author}{\bibinfo{person}{Amil Merchant}, \bibinfo{person}{Simon
  Batzner}, \bibinfo{person}{Samuel~S. Schoenholz}, \bibinfo{person}{Muratahan
  Aykol}, \bibinfo{person}{Gowoon Cheon}, {and} \bibinfo{person}{Ekin~Dogus
  Cubuk}.} \bibinfo{year}{2023}\natexlab{}.
\newblock \showarticletitle{Scaling deep learning for materials discovery}.
\newblock \bibinfo{journal}{\emph{Nature}} \bibinfo{volume}{624},
  \bibinfo{number}{7990} (\bibinfo{year}{2023}), \bibinfo{pages}{80–85}.
\newblock
\showISSN{1476-4687}


\bibitem[Merkwirth and Lengauer(2005)]%
        {Merkwirth2005Automatic}
\bibfield{author}{\bibinfo{person}{Christian Merkwirth} {and}
  \bibinfo{person}{Thomas Lengauer}.} \bibinfo{year}{2005}\natexlab{}.
\newblock \showarticletitle{Automatic Generation of Complementary Descriptors
  with Molecular Graph Networks}.
\newblock \bibinfo{journal}{\emph{Journal of Chemical Information and
  Modeling}} \bibinfo{volume}{45}, \bibinfo{number}{5} (\bibinfo{year}{2005}),
  \bibinfo{pages}{1159–1168}.
\newblock
\showISSN{1549-960X}


\bibitem[Merkys et~al\mbox{.}(2016)]%
        {Merkys2016COD}
\bibfield{author}{\bibinfo{person}{Andrius Merkys}, \bibinfo{person}{Antanas
  Vaitkus}, \bibinfo{person}{Justas Butkus}, \bibinfo{person}{Mykolas
  Okulič-Kazarinas}, \bibinfo{person}{Visvaldas Kairys}, {and}
  \bibinfo{person}{Saulius Gražulis}.} \bibinfo{year}{2016}\natexlab{}.
\newblock \showarticletitle{COD::CIF::Parser: an error-correcting CIF parser
  for the Perl language}.
\newblock \bibinfo{journal}{\emph{Journal of Applied Crystallography}}
  \bibinfo{volume}{49}, \bibinfo{number}{1} (\bibinfo{year}{2016}),
  \bibinfo{pages}{292–301}.
\newblock
\showISSN{1600-5767}


\bibitem[Merkys et~al\mbox{.}(2023)]%
        {Merkys2023Graph}
\bibfield{author}{\bibinfo{person}{Andrius Merkys}, \bibinfo{person}{Antanas
  Vaitkus}, \bibinfo{person}{Algirdas Grybauskas}, \bibinfo{person}{Aleksandras
  Konovalovas}, \bibinfo{person}{Miguel Quirós}, {and}
  \bibinfo{person}{Saulius Gražulis}.} \bibinfo{year}{2023}\natexlab{}.
\newblock \showarticletitle{Graph isomorphism-based algorithm for
  cross-checking chemical and crystallographic descriptions}.
\newblock \bibinfo{journal}{\emph{Journal of Cheminformatics}}
  \bibinfo{volume}{15}, \bibinfo{number}{1} (\bibinfo{year}{2023}).
\newblock
\showISSN{1758-2946}


\bibitem[Momma and Izumi(2008)]%
        {Momma2008VESTA}
\bibfield{author}{\bibinfo{person}{Koichi Momma} {and} \bibinfo{person}{Fujio
  Izumi}.} \bibinfo{year}{2008}\natexlab{}.
\newblock \showarticletitle{VESTA: a three-dimensional visualization system for
  electronic and structural analysis}.
\newblock \bibinfo{journal}{\emph{Journal of Applied Crystallography}}
  \bibinfo{volume}{41}, \bibinfo{number}{3} (\bibinfo{year}{2008}),
  \bibinfo{pages}{653–658}.
\newblock
\showISSN{0021-8898}


\bibitem[Paszke et~al\mbox{.}(2019)]%
        {Paszke2019PyTorch}
\bibfield{author}{\bibinfo{person}{Adam Paszke}, \bibinfo{person}{Sam Gross},
  \bibinfo{person}{Francisco Massa}, \bibinfo{person}{Adam Lerer},
  \bibinfo{person}{James Bradbury}, \bibinfo{person}{Gregory Chanan},
  \bibinfo{person}{Trevor Killeen}, \bibinfo{person}{Zeming Lin},
  \bibinfo{person}{Natalia Gimelshein}, \bibinfo{person}{Luca Antiga},
  \bibinfo{person}{Alban Desmaison}, \bibinfo{person}{Andreas Kopf},
  \bibinfo{person}{Edward Yang}, \bibinfo{person}{Zachary DeVito},
  \bibinfo{person}{Martin Raison}, \bibinfo{person}{Alykhan Tejani},
  \bibinfo{person}{Sasank Chilamkurthy}, \bibinfo{person}{Benoit Steiner},
  \bibinfo{person}{Lu Fang}, \bibinfo{person}{Junjie Bai}, {and}
  \bibinfo{person}{Soumith Chintala}.} \bibinfo{year}{2019}\natexlab{}.
\newblock \showarticletitle{PyTorch: An Imperative Style, High-Performance Deep
  Learning Library}. In \bibinfo{booktitle}{\emph{Advances in Neural
  Information Processing Systems}}, Vol.~\bibinfo{volume}{32}.
\newblock


\bibitem[Quirós et~al\mbox{.}(2018)]%
        {Quirs2018Using}
\bibfield{author}{\bibinfo{person}{Miguel Quirós}, \bibinfo{person}{Saulius
  Gražulis}, \bibinfo{person}{Saulė Girdzijauskaitė},
  \bibinfo{person}{Andrius Merkys}, {and} \bibinfo{person}{Antanas Vaitkus}.}
  \bibinfo{year}{2018}\natexlab{}.
\newblock \showarticletitle{Using SMILES strings for the description of
  chemical connectivity in the Crystallography Open Database}.
\newblock \bibinfo{journal}{\emph{Journal of Cheminformatics}}
  \bibinfo{volume}{10}, \bibinfo{number}{1} (\bibinfo{year}{2018}).
\newblock
\showISSN{1758-2946}


\bibitem[Ramakrishnan et~al\mbox{.}(2014)]%
        {Ramakrishnan2014Quantum}
\bibfield{author}{\bibinfo{person}{Raghunathan Ramakrishnan},
  \bibinfo{person}{Pavlo~O. Dral}, \bibinfo{person}{Matthias Rupp}, {and}
  \bibinfo{person}{O.~Anatole von Lilienfeld}.}
  \bibinfo{year}{2014}\natexlab{}.
\newblock \showarticletitle{Quantum chemistry structures and properties of 134
  kilo molecules}.
\newblock \bibinfo{journal}{\emph{Scientific Data}} \bibinfo{volume}{1},
  \bibinfo{number}{1} (\bibinfo{year}{2014}).
\newblock
\showISSN{2052-4463}


\bibitem[Ruddigkeit et~al\mbox{.}(2012)]%
        {Ruddigkeit2012Enumeration}
\bibfield{author}{\bibinfo{person}{Lars Ruddigkeit}, \bibinfo{person}{Ruud van
  Deursen}, \bibinfo{person}{Lorenz~C. Blum}, {and} \bibinfo{person}{Jean-Louis
  Reymond}.} \bibinfo{year}{2012}\natexlab{}.
\newblock \showarticletitle{Enumeration of 166 Billion Organic Small Molecules
  in the Chemical Universe Database GDB-17}.
\newblock \bibinfo{journal}{\emph{Journal of Chemical Information and
  Modeling}} \bibinfo{volume}{52}, \bibinfo{number}{11} (\bibinfo{year}{2012}),
  \bibinfo{pages}{2864–2875}.
\newblock
\showISSN{1549-960X}


\bibitem[Scarselli et~al\mbox{.}(2009)]%
        {Scarselli2009Graph}
\bibfield{author}{\bibinfo{person}{F. Scarselli}, \bibinfo{person}{M. Gori},
  \bibinfo{person}{Ah~Chung Tsoi}, \bibinfo{person}{M. Hagenbuchner}, {and}
  \bibinfo{person}{G. Monfardini}.} \bibinfo{year}{2009}\natexlab{}.
\newblock \showarticletitle{The Graph Neural Network Model}.
\newblock \bibinfo{journal}{\emph{IEEE Transactions on Neural Networks}}
  \bibinfo{volume}{20}, \bibinfo{number}{1} (\bibinfo{year}{2009}),
  \bibinfo{pages}{61–80}.
\newblock
\showISSN{1941-0093}


\bibitem[Schmidhuber(2015)]%
        {Schmidhuber2015Deep}
\bibfield{author}{\bibinfo{person}{J\"{u}rgen Schmidhuber}.}
  \bibinfo{year}{2015}\natexlab{}.
\newblock \showarticletitle{Deep learning in neural networks: An overview}.
\newblock \bibinfo{journal}{\emph{Neural Networks}}  \bibinfo{volume}{61}
  (\bibinfo{year}{2015}), \bibinfo{pages}{85–117}.
\newblock
\showISSN{0893-6080}


\bibitem[Schmidt et~al\mbox{.}(2021)]%
        {Schmidt2021Crystal}
\bibfield{author}{\bibinfo{person}{Jonathan Schmidt}, \bibinfo{person}{Love
  Pettersson}, \bibinfo{person}{Claudio Verdozzi}, \bibinfo{person}{Silvana
  Botti}, {and} \bibinfo{person}{Miguel A.~L. Marques}.}
  \bibinfo{year}{2021}\natexlab{}.
\newblock \showarticletitle{Crystal graph attention networks for the prediction
  of stable materials}.
\newblock \bibinfo{journal}{\emph{Science Advances}} \bibinfo{volume}{7},
  \bibinfo{number}{49} (\bibinfo{year}{2021}).
\newblock
\showISSN{2375-2548}


\bibitem[Sch{\"u}tt et~al\mbox{.}(2017)]%
        {Schutt2017SchNet}
\bibfield{author}{\bibinfo{person}{Kristof Sch{\"u}tt},
  \bibinfo{person}{Pieter-Jan Kindermans}, \bibinfo{person}{Huziel~Enoc
  Sauceda~Felix}, \bibinfo{person}{Stefan Chmiela}, \bibinfo{person}{Alexandre
  Tkatchenko}, {and} \bibinfo{person}{Klaus-Robert M{\"u}ller}.}
  \bibinfo{year}{2017}\natexlab{}.
\newblock \showarticletitle{Schnet: A continuous-filter convolutional neural
  network for modeling quantum interactions}.
\newblock \bibinfo{journal}{\emph{Advances in Neural Information Processing
  Systems (NeurIPS)}} (\bibinfo{year}{2017}).
\newblock


\bibitem[Slater(1964)]%
        {Slater1964Atomic}
\bibfield{author}{\bibinfo{person}{J.~C. Slater}.}
  \bibinfo{year}{1964}\natexlab{}.
\newblock \showarticletitle{Atomic Radii in Crystals}.
\newblock \bibinfo{journal}{\emph{The Journal of Chemical Physics}}
  \bibinfo{volume}{41}, \bibinfo{number}{10} (\bibinfo{year}{1964}),
  \bibinfo{pages}{3199–3204}.
\newblock
\showISSN{1089-7690}


\bibitem[Sterling and Irwin(2015)]%
        {Sterling2015ZINC}
\bibfield{author}{\bibinfo{person}{Teague Sterling} {and}
  \bibinfo{person}{John~J. Irwin}.} \bibinfo{year}{2015}\natexlab{}.
\newblock \showarticletitle{ZINC 15 – Ligand Discovery for Everyone}.
\newblock \bibinfo{journal}{\emph{Journal of Chemical Information and
  Modeling}} \bibinfo{volume}{55}, \bibinfo{number}{11} (\bibinfo{year}{2015}),
  \bibinfo{pages}{2324–2337}.
\newblock
\showISSN{1549-960X}


\bibitem[Thanapalasingam et~al\mbox{.}(2023)]%
        {Thanapalasingam2023IntelliGraphs}
\bibfield{author}{\bibinfo{person}{Thiviyan Thanapalasingam},
  \bibinfo{person}{Emile van Krieken}, \bibinfo{person}{Peter Bloem}, {and}
  \bibinfo{person}{Paul Groth}.} \bibinfo{year}{2023}\natexlab{}.
\newblock \showarticletitle{IntelliGraphs: Datasets for Benchmarking Knowledge
  Graph Generation}.
\newblock  (\bibinfo{year}{2023}).
\newblock


\bibitem[Tran et~al\mbox{.}(2023)]%
        {Tran2023Open}
\bibfield{author}{\bibinfo{person}{Richard Tran}, \bibinfo{person}{Janice Lan},
  \bibinfo{person}{Muhammed Shuaibi}, \bibinfo{person}{Brandon~M. Wood},
  \bibinfo{person}{Siddharth Goyal}, \bibinfo{person}{Abhishek Das},
  \bibinfo{person}{Javier Heras-Domingo}, \bibinfo{person}{Adeesh Kolluru},
  \bibinfo{person}{Ammar Rizvi}, \bibinfo{person}{Nima Shoghi},
  \bibinfo{person}{Anuroop Sriram}, \bibinfo{person}{Félix Therrien},
  \bibinfo{person}{Jehad Abed}, \bibinfo{person}{Oleksandr Voznyy},
  \bibinfo{person}{Edward~H. Sargent}, \bibinfo{person}{Zachary Ulissi}, {and}
  \bibinfo{person}{C.~Lawrence Zitnick}.} \bibinfo{year}{2023}\natexlab{}.
\newblock \showarticletitle{The Open Catalyst 2022 (OC22) Dataset and
  Challenges for Oxide Electrocatalysts}.
\newblock \bibinfo{journal}{\emph{ACS Catalysis}} \bibinfo{volume}{13},
  \bibinfo{number}{5} (\bibinfo{year}{2023}), \bibinfo{pages}{3066–3084}.
\newblock
\showISSN{2155-5435}


\bibitem[Vaitkus et~al\mbox{.}(2021)]%
        {Vaitkus2021Validation}
\bibfield{author}{\bibinfo{person}{Antanas Vaitkus}, \bibinfo{person}{Andrius
  Merkys}, {and} \bibinfo{person}{Saulius Gražulis}.}
  \bibinfo{year}{2021}\natexlab{}.
\newblock \showarticletitle{Validation of the Crystallography Open Database
  using the Crystallographic Information Framework}.
\newblock \bibinfo{journal}{\emph{Journal of Applied Crystallography}}
  \bibinfo{volume}{54}, \bibinfo{number}{2} (\bibinfo{year}{2021}),
  \bibinfo{pages}{661–672}.
\newblock
\showISSN{1600-5767}


\bibitem[Vaitkus et~al\mbox{.}(2023)]%
        {Vaitkus2023workflow}
\bibfield{author}{\bibinfo{person}{Antanas Vaitkus}, \bibinfo{person}{Andrius
  Merkys}, \bibinfo{person}{Thomas Sander}, \bibinfo{person}{Miguel Quirós},
  \bibinfo{person}{Paul~A. Thiessen}, \bibinfo{person}{Evan~E. Bolton}, {and}
  \bibinfo{person}{Saulius Gražulis}.} \bibinfo{year}{2023}\natexlab{}.
\newblock \showarticletitle{A workflow for deriving chemical entities from
  crystallographic data and its application to the Crystallography Open
  Database}.
\newblock \bibinfo{journal}{\emph{Journal of Cheminformatics}}
  \bibinfo{volume}{15}, \bibinfo{number}{1} (\bibinfo{year}{2023}).
\newblock
\showISSN{1758-2946}


\bibitem[Veli{\v{c}}kovi{\'c} et~al\mbox{.}(2018)]%
        {velivckovic2018graph}
\bibfield{author}{\bibinfo{person}{Petar Veli{\v{c}}kovi{\'c}},
  \bibinfo{person}{Guillem Cucurull}, \bibinfo{person}{Arantxa Casanova},
  \bibinfo{person}{Adriana Romero}, \bibinfo{person}{Pietro Li{\`o}}, {and}
  \bibinfo{person}{Yoshua Bengio}.} \bibinfo{year}{2018}\natexlab{}.
\newblock \showarticletitle{Graph Attention Networks}. In
  \bibinfo{booktitle}{\emph{International Conference on Learning
  Representations (ICLR)}}.
\newblock


\bibitem[Vignac et~al\mbox{.}(2023)]%
        {vignac2023digress}
\bibfield{author}{\bibinfo{person}{Clement Vignac}, \bibinfo{person}{Igor
  Krawczuk}, \bibinfo{person}{Antoine Siraudin}, \bibinfo{person}{Bohan Wang},
  \bibinfo{person}{Volkan Cevher}, {and} \bibinfo{person}{Pascal Frossard}.}
  \bibinfo{year}{2023}\natexlab{}.
\newblock \showarticletitle{DiGress: Discrete Denoising diffusion for graph
  generation}. In \bibinfo{booktitle}{\emph{The Eleventh International
  Conference on Learning Representations}}.
\newblock


\bibitem[Wang et~al\mbox{.}(2023)]%
        {Wang2023Graph}
\bibfield{author}{\bibinfo{person}{Yuyang Wang}, \bibinfo{person}{Zijie Li},
  {and} \bibinfo{person}{Amir Barati~Farimani}.}
  \bibinfo{year}{2023}\natexlab{}.
\newblock \bibinfo{booktitle}{\emph{Graph Neural Networks for Molecules}}.
\newblock \bibinfo{pages}{21–66}.
\newblock
\showISBNx{9783031371967}
\showISSN{2542-4483}


\bibitem[Wang et~al\mbox{.}(2019)]%
        {Wang2018Dynamic}
\bibfield{author}{\bibinfo{person}{Yue Wang}, \bibinfo{person}{Yongbin Sun},
  \bibinfo{person}{Ziwei Liu}, \bibinfo{person}{Sanjay~E Sarma},
  \bibinfo{person}{Michael~M Bronstein}, {and} \bibinfo{person}{Justin~M
  Solomon}.} \bibinfo{year}{2019}\natexlab{}.
\newblock \showarticletitle{{Dynamic graph CNN for learning on point clouds}}.
\newblock \bibinfo{journal}{\emph{ACM Transactions on Graphics (tog)}}
  (\bibinfo{year}{2019}).
\newblock


\bibitem[Ward et~al\mbox{.}(2016)]%
        {Ward2016general}
\bibfield{author}{\bibinfo{person}{Logan Ward}, \bibinfo{person}{Ankit
  Agrawal}, \bibinfo{person}{Alok Choudhary}, {and}
  \bibinfo{person}{Christopher Wolverton}.} \bibinfo{year}{2016}\natexlab{}.
\newblock \showarticletitle{A general-purpose machine learning framework for
  predicting properties of inorganic materials}.
\newblock \bibinfo{journal}{\emph{npj Computational Materials}}
  \bibinfo{volume}{2}, \bibinfo{number}{1} (\bibinfo{year}{2016}).
\newblock
\showISSN{2057-3960}


\bibitem[West(2022)]%
        {West2022Solid}
\bibfield{author}{\bibinfo{person}{Anthony~R West}.}
  \bibinfo{year}{2022}\natexlab{}.
\newblock \bibinfo{booktitle}{\emph{Solid State Chemistry and its Applications}
  (\bibinfo{edition}{2} ed.)}.
\newblock \bibinfo{publisher}{John Wiley \& Sons}.
\newblock


\bibitem[Wu et~al\mbox{.}(2018)]%
        {Wu2017MoleculeNet}
\bibfield{author}{\bibinfo{person}{Zhenqin Wu}, \bibinfo{person}{Bharath
  Ramsundar}, \bibinfo{person}{Evan~N Feinberg}, \bibinfo{person}{Joseph
  Gomes}, \bibinfo{person}{Caleb Geniesse}, \bibinfo{person}{Aneesh~S Pappu},
  \bibinfo{person}{Karl Leswing}, {and} \bibinfo{person}{Vijay Pande}.}
  \bibinfo{year}{2018}\natexlab{}.
\newblock \showarticletitle{MoleculeNet: a benchmark for molecular machine
  learning}.
\newblock \bibinfo{journal}{\emph{Chemical science}} (\bibinfo{year}{2018}).
\newblock


\bibitem[Xie et~al\mbox{.}(2022)]%
        {Xie2021Crystal}
\bibfield{author}{\bibinfo{person}{Tian Xie}, \bibinfo{person}{Xiang Fu},
  \bibinfo{person}{Octavian-Eugen Ganea}, \bibinfo{person}{Regina Barzilay},
  {and} \bibinfo{person}{Tommi~S. Jaakkola}.} \bibinfo{year}{2022}\natexlab{}.
\newblock \showarticletitle{Crystal Diffusion Variational Autoencoder for
  Periodic Material Generation}. In \bibinfo{booktitle}{\emph{International
  Conference on Learning Representations (ICLR)}}.
\newblock


\bibitem[Xie and Grossman(2018)]%
        {Xie2018Crystal}
\bibfield{author}{\bibinfo{person}{Tian Xie} {and} \bibinfo{person}{Jeffrey~C.
  Grossman}.} \bibinfo{year}{2018}\natexlab{}.
\newblock \showarticletitle{Crystal Graph Convolutional Neural Networks for an
  Accurate and Interpretable Prediction of Material Properties}.
\newblock \bibinfo{journal}{\emph{Physical Review Letters}}
  \bibinfo{volume}{120}, \bibinfo{number}{14} (\bibinfo{year}{2018}).
\newblock
\showISSN{1079-7114}


\bibitem[Xu et~al\mbox{.}(2019)]%
        {Xu2018How}
\bibfield{author}{\bibinfo{person}{Keyulu Xu}, \bibinfo{person}{Weihua Hu},
  \bibinfo{person}{Jure Leskovec}, {and} \bibinfo{person}{Stefanie Jegelka}.}
  \bibinfo{year}{2019}\natexlab{}.
\newblock \showarticletitle{How Powerful are Graph Neural Networks?}. In
  \bibinfo{booktitle}{\emph{International Conference on Learning
  Representations (ICLR)}}.
\newblock


\bibitem[Yang et~al\mbox{.}(2023)]%
        {Yang2022Graph}
\bibfield{author}{\bibinfo{person}{Chenxiao Yang}, \bibinfo{person}{Qitian Wu},
  \bibinfo{person}{Jiahua Wang}, {and} \bibinfo{person}{Junchi Yan}.}
  \bibinfo{year}{2023}\natexlab{}.
\newblock \showarticletitle{Graph Neural Networks are Inherently Good
  Generalizers: Insights by Bridging {GNN}s and {MLP}s}. In
  \bibinfo{booktitle}{\emph{International Conference on Learning
  Representations (ICLR)}}.
\newblock


\bibitem[Yang et~al\mbox{.}(2022)]%
        {Yang2022Big}
\bibfield{author}{\bibinfo{person}{Ruo~Xi Yang}, \bibinfo{person}{Caitlin~A.
  McCandler}, \bibinfo{person}{Oxana Andriuc}, \bibinfo{person}{Martin Siron},
  \bibinfo{person}{Rachel Woods-Robinson}, \bibinfo{person}{Matthew~K. Horton},
  {and} \bibinfo{person}{Kristin~A. Persson}.} \bibinfo{year}{2022}\natexlab{}.
\newblock \showarticletitle{Big Data in a Nano World: A Review on
  Computational, Data-Driven Design of Nanomaterials Structures, Properties,
  and Synthesis}.
\newblock \bibinfo{journal}{\emph{ACS Nano}} \bibinfo{volume}{16},
  \bibinfo{number}{12} (\bibinfo{year}{2022}), \bibinfo{pages}{19873–19891}.
\newblock
\showISSN{1936-086X}


\bibitem[Zagorac et~al\mbox{.}(2019)]%
        {Zagorac2019Recent}
\bibfield{author}{\bibinfo{person}{D. Zagorac}, \bibinfo{person}{H.
  M\"{u}ller}, \bibinfo{person}{S. Ruehl}, \bibinfo{person}{J. Zagorac}, {and}
  \bibinfo{person}{S. Rehme}.} \bibinfo{year}{2019}\natexlab{}.
\newblock \showarticletitle{Recent developments in the Inorganic Crystal
  Structure Database: theoretical crystal structure data and related features}.
\newblock \bibinfo{journal}{\emph{Journal of Applied Crystallography}}
  \bibinfo{volume}{52}, \bibinfo{number}{5} (\bibinfo{year}{2019}),
  \bibinfo{pages}{918–925}.
\newblock
\showISSN{1600-5767}


\bibitem[Zeni et~al\mbox{.}(2023)]%
        {Zeni2023MatterGen}
\bibfield{author}{\bibinfo{person}{Claudio Zeni}, \bibinfo{person}{Robert
  Pinsler}, \bibinfo{person}{Daniel Z\"{u}gner}, \bibinfo{person}{Andrew
  Fowler}, \bibinfo{person}{Matthew Horton}, \bibinfo{person}{Xiang Fu},
  \bibinfo{person}{Sasha Shysheya}, \bibinfo{person}{Jonathan Crabbé},
  \bibinfo{person}{Lixin Sun}, \bibinfo{person}{Jake Smith},
  \bibinfo{person}{Bichlien Nguyen}, \bibinfo{person}{Hannes Schulz},
  \bibinfo{person}{Sarah Lewis}, \bibinfo{person}{Chin-Wei Huang},
  \bibinfo{person}{Ziheng Lu}, \bibinfo{person}{Yichi Zhou},
  \bibinfo{person}{Han Yang}, \bibinfo{person}{Hongxia Hao},
  \bibinfo{person}{Jielan Li}, \bibinfo{person}{Ryota Tomioka}, {and}
  \bibinfo{person}{Tian Xie}.} \bibinfo{year}{2023}\natexlab{}.
\newblock \showarticletitle{MatterGen: a generative model for inorganic
  materials design}.
\newblock  (\bibinfo{year}{2023}).
\newblock


\bibitem[Zitnik and Leskovec(2017)]%
        {Zitnik2017Predicting}
\bibfield{author}{\bibinfo{person}{Marinka Zitnik} {and} \bibinfo{person}{Jure
  Leskovec}.} \bibinfo{year}{2017}\natexlab{}.
\newblock \showarticletitle{Predicting multicellular function through
  multi-layer tissue networks}.
\newblock \bibinfo{journal}{\emph{Bioinformatics}} \bibinfo{volume}{33},
  \bibinfo{number}{14} (\bibinfo{year}{2017}), \bibinfo{pages}{i190–i198}.
\newblock
\showISSN{1367-4811}


\end{thebibliography}
\clearpage
\appendix

\section{Data generation}
\subsection{CIF construction}\label{app:cif_gen}
We construct CIFs according to the crystal types described in West et al.~\cite{West2022Solid} using the \texttt{ase.spacegroup.crystal} function from the atomic simulation environment (ASE)~\cite{HjorthLarsen2017Atomic}. Each crystal type is described by its fractional atomic positions and its spacegroup. West et al.~\cite{West2022Solid} also provides table values for the unit cell parameters of select materials for each crystal type, which we use to estimate the unit cell parameters of each combination of elements and crystal types. This is done by fitting linear functions to the relationship between each unit cell parameter and the sum of atomic radii of the elements in the chemical formula. The CIF generation code can be found \href{https://github.com/UlrikFriisJensen/CHILI/blob/main/generation/generate_cifs.py}{\underline{here}}.

\subsection{Crystallography Open Database query}\label{app:chili100k}
The CIFs from the Crystallography Open Database (COD)~\cite{Downs2003TheAM, Graulis2009Crystallography, Graulis2011Crystallography, Graulis2015Computing, Merkys2016COD, Quirs2018Using, Vaitkus2021Validation, Merkys2023Graph, Vaitkus2023workflow} was queried using the code found \href{https://github.com/UlrikFriisJensen/CHILI/blob/main/generation/database_query.py}{\underline{here}}. The CIFs were then cleaned using the code found \href{https://github.com/UlrikFriisJensen/CHILI/blob/main/generation/cif_cleaning.py}{\underline{here}}. The cleaning process is described in words below.

We used the {\tt ase.io.read} function from ASE~\cite{HjorthLarsen2017Atomic} to load the CIFs. Major issues that were known and could be fixed included: generation loop errors, empty columns which we fill with 0's, and syntax errors - specifically non-closed parenthesis to indicate uncertainties. Issues such as parts of the file missing or less commonly seen errors, could not be fixed and the files were therefore removed from the dataset. 

Occasionally, we would observe minor issues where the periodicity of the unit cell would be affected by an insufficient floating-point precision on certain rational fractions such as 1/3 or 1/9. Through an empirical study, we found that a precision of less that five decimal places would lead to periodicity errors, potentially displacing atoms out of the intended symmetry site. We detected such floating-point numbers and increased their precision to five decimal places. 

We also found that some CIFs contain two-letter elemental symbols that have both letters capitalized. This leads to issues where the elemental symbols are read incorrectly, such as copper (Cu) being read as carbon (C). This is difficult to detect, but to our knowledge quite rare. We handled the worst cases, where a material is interpreted as containing no metallic elements, by simply removing these files from the dataset.

\subsection{Nanomaterial graph generation}\label{app:nanoMaterialGraph}
We here in more detail explain the graph generation of the \texttt{CHILI} datasets, which is outlined in Figure \ref{fig:generation_overview} (step 3 and 4). The nanoparticle generation is done using the \texttt{generate\_nanoparticles} utility function from DebyeCalculator~\cite{Johansen2024GPU} (\href{https://github.com/FrederikLizakJohansen/DebyeCalculator/blob/main/debyecalculator/utility/generate.py#L96-L364}{Found here}). This approach relies on approximating the real nanoparticle structure as cutouts from crystalline materials. 

\textbf{Step 3:} After reading the CIF with ASE~\cite{HjorthLarsen2017Atomic} (step 2), the unit cell is expanded into a supercell that we ensure extends at least the largest nanoparticle diameter plus an additional 5 Ångströms (Å) of padding along each axis. The supercell is then centered by moving the most central metal atom into origo and the distance of all atoms to the center is calculated to make it easier to determine which atoms are within a cutoff distance. 

The atom connectivity in the supercell is then determined by finding the overlaps between each atoms estimated interaction neighborhood. We define the interaction neighborhoods as 125 \% of the elements atomic radii in crystals. Atomic radii values was accessed using the Mendeleev package~\cite{Mentel2023Mendeleev} (Data from Slater et al.~\cite{Slater1964Atomic}). The extra 25 \% is added to allow for atoms in distorted coordination environments to still be connected. In the border-case where all atoms are too far apart and no overlaps between interaction neighborhoods are found, 110 \% of the smallest distance in the unit cell is used as the radius of the interaction neighborhood. This approach to determine atom connectivity creates more edges in the final nanomaterial graph than using the true 1st coordination shell of the atoms. However, that would be way more expensive to calculate and we think this approach is a more realistic approximation of interaction than using a fixed-size local neighborhood to determine edges or using a fully connected graph.

\textbf{Step 4:} The metallic core of the nanoparticle is found simply by checking whether the distance of a metal atom from origo is smaller than or equal to the given radius. The non-metals in the nanoparticle is determined by one of two things: (1) if the distance to origo is less than or equal to the given radius or (2) if its interaction neighborhood is overlapping with the metallic core of the nanoparticle. For both datasets we generate nanoparticles with 5, 10, 15, 20 and 25 Å as the radii. 

\subsection{Scattering data simulation}\label{app:scatteringSimulation}
The simulation of scattering data, step 5 in Figure \ref{fig:generation_overview}, in this paper was done using DebyeCalculator~\cite{Johansen2024GPU}. 
\texttt{saxs} and \texttt{sans} was calculated using the \texttt{iq} function with the parameters shown in Table \ref{tab:debye_parameters}. \texttt{xrd}, \texttt{nd}, \texttt{xPDF} and \texttt{nPDF} was calculated using the \texttt{\_get\_all} function with the parameters shown in Table \ref{tab:debye_parameters}.

\begin{table}
    \caption{DebyeCalculator parameters used for simulation of scattering data.}
    \label{tab:debye_parameters}
    \small
    \begin{tabular}{lccc}
        \toprule
        Parameter & \texttt{saxs}/\texttt{sans} & \texttt{xrd}/\texttt{nd} & \texttt{xPDF}/\texttt{nPDF}\\
        \midrule
        \texttt{qmin} & 0 & 1 & 1 \\
        \texttt{qmax} & 3 & 30 & 30 \\
        \texttt{qstep} & 0.01 & 0.05 & 0.05 \\
        \texttt{biso} & 0.3 & 0.3 & 0.3 \\
        \texttt{rmin} & --- & --- & 0 \\
        \texttt{rmax} & --- & --- & 60 \\
        \texttt{rstep} & --- & --- & 0.01 \\
        \bottomrule
    \end{tabular}
\end{table}

\section{More statistics}\label{app:statistics}
Here we present more statistics for the \texttt{CHILI} datasets that were not essential for the main paper. 
\subsection{\texttt{CHILI-3K} crystal type distribution}
\begin{figure}[!htb]
    \centering
    \includegraphics[width=\columnwidth]{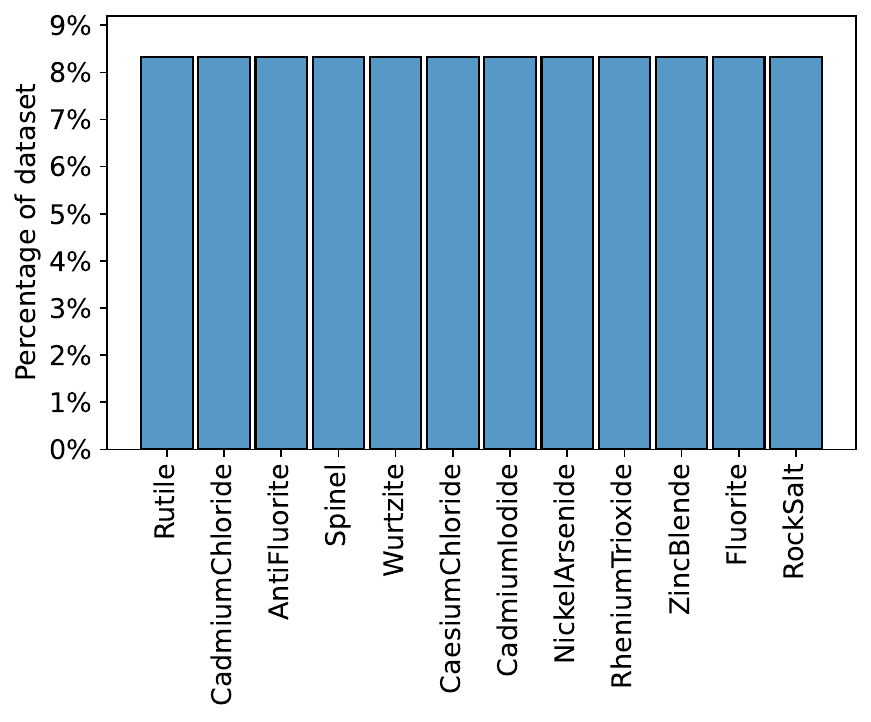}
    \caption{Distribution of crystal types in the \texttt{CHILI-3K} dataset.}
    \label{fig:dist_crystal_types}
\end{figure}

The distribution of crystal types in the \texttt{CHILI-3K} dataset is shown in Figure \ref{fig:dist_crystal_types} to verify that an equal amount of each crystal type was simulated.

\subsection{\texttt{CHILI-100K} subset}
\begin{figure}[!htb]
    \centering
    \includegraphics[width=\columnwidth]{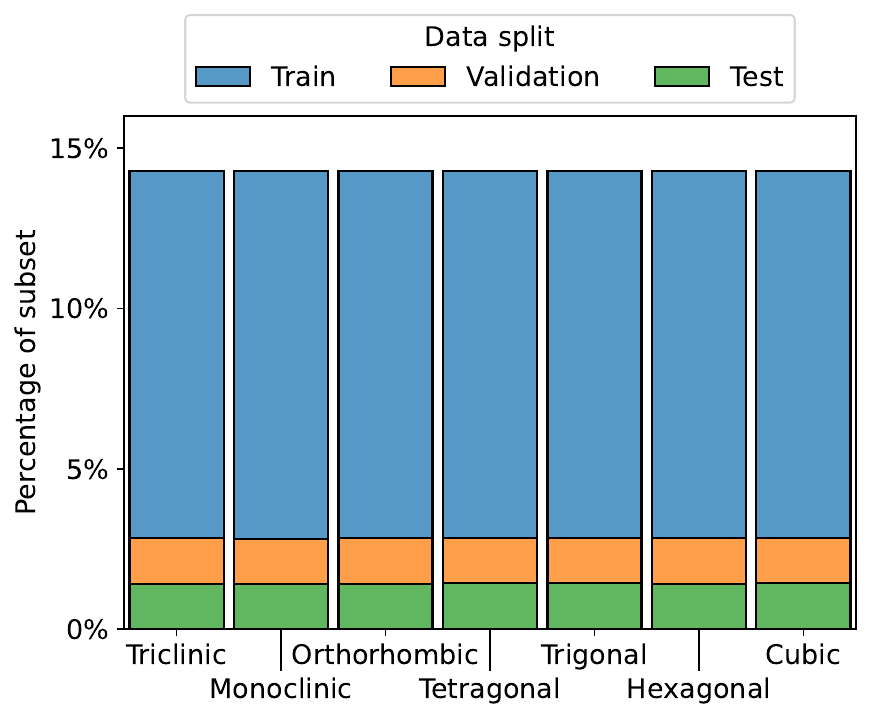}
    \caption{Distribution of crystal systems in the \texttt{CHILI-100K} subset.}
    \label{fig:dist_crystal_system_100k_subset}
\end{figure}

For benchmarking we use a stratified subset of the \texttt{CHILI-100K} dataset with 425 samples from each crystal type.
The distribution of crystal systems in the \texttt{CHILI-100K} subset is shown in Figure \ref{fig:dist_crystal_system_100k_subset}. We see that the crystal systems are indeed equally distributed in the subset and each data split.

\begin{figure}[!htb]
    \centering
    \includegraphics[width=\columnwidth]{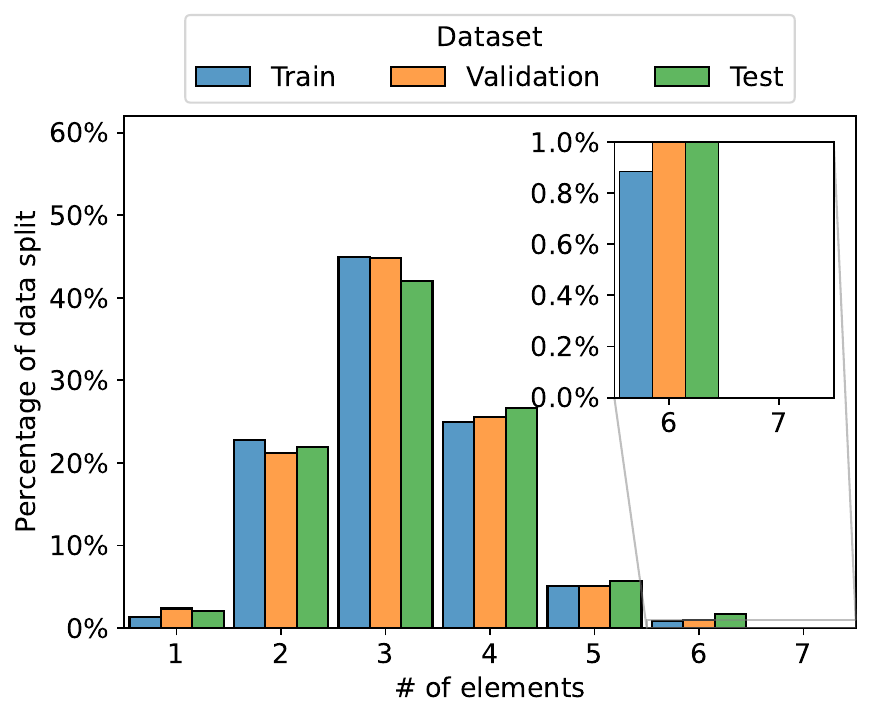}
    \caption{Distribution of number of elements in the \texttt{CHILI-100K} subset. The inset plot shows 6 and 7 elements at a more appropiate y-axis scale.}
    \label{fig:dist_nElements_100k_subset}
\end{figure}

The number of elements in the \texttt{CHILI-100K} subset is shown in Figure \ref{fig:dist_nElements_100k_subset}. We see that the distribution of number of elements match across the data splits, even though the stratified strategy did not consider this.

\section{Model- and task setup}\label{app:modeltasksetup}

\subsection{Model setup}

\textbf{GCN:} We use the Graph Convolutional Network (GCN)~\cite{Kipf2016Semi} as our baseline benchmarking model. Using the hyperparameters described in the experimental section of the original paper, we train a two-layer GCN with a hidden layer size of 32 using the Adam optimizer~\cite{Kingma2014Adam} with a learning rate of 0.01~\cite{Kipf2016Semi}. 

\textbf{GraphSAGE:} We use the GraphSAGE (SAmple and aggreGatE) ~\cite{Hamilton2017Inductive} model with the default mean aggregator. The same hyperparameters are used for the GraphSAGE model as we used for the GCN model.

\textbf{GAT:} We use the graph attention network (GAT)~\cite{velivckovic2018graph} model with the GATConv convolution. As described in the paper, we use a hidden layer size of 64 and then use our default values from the GCN model for the rest.

\textbf{GIN:} We use the graph isomorphism network (GIN)~\cite{Xu2018How} model with $\epsilon$ fixed to 0, the GIN-0 variant. The same hyperparameters are used for the GIN model as we used for the GCN model.

\textbf{EdgeCNN:} We use the EdgeCNN~\cite{Wang2018Dynamic} model, but without recomputing the graph at every layer as it is not a part of the out-of-the-box model. As described in the paper, we use a 4-layer EdgeCNN model with a hidden layer size of 64. The rest of the hyperparameters use the default values from the GCN model. 

\textbf{GraphUNet:} We use the graph U-Net~\cite{Gao2019Graph} model with the default pooling ratio of 0.5. The paper observes improved performance with deeper networks until a depth of 4, however we use only a depth of 2 because of memory constraints. The other hyperparameters are the same as the default values from the GCN model.

\textbf{PMLP:} We use the propagational multi-layer perceptron (PMLP) ~\cite{Yang2022Graph} model with default initialization values. The same hyperparameters are used for the PMLP model as we used for the GCN model.

\textbf{MLP:} We use a simple 4-layer multilayered perception (MLP) model with a hidden layer size of 128. A ReLU activation function is applied following every layer except the final one. 

\subsection{Model complexity}
We measure the complexity of the off-the-shelf GNN models by their number of trainable parameters. The number of trainable parameters change for the node-level tasks, \texttt{atomic\_number} classification and \texttt{pos\_abs} regression, because the feature dimensions of the final GNN layer are different. The number of trainable parameters for the models are shown in Table \ref{app:num_param}.
\begin{table}
    \caption{Number of trainable parameters in the GNN backbone. These numbers are not accurate for the node-level tasks, \texttt{atomic\_number} classification and \texttt{pos\_abs} regression, because the feature dimensions of the final GNN layer are different.}
    \label{app:num_param}
\begin{tabular}{lr}
\toprule
 Model & Trainable parameters \\
\midrule
GCN & 2368 \\
PMLP & 2368 \\
GraphSAGE & 4640 \\
GAT & 4928 \\
GraphUNet & 5600 \\
GIN & 7584 \\
EdgeCNN & 42368 \\
\bottomrule
\end{tabular}
\end{table}

\subsection{Task setup}
For the benchmarking of prediction tasks, we chose all classification tasks, see Table \ref{tab:tasks_overview}, and all regression tasks except for SANS, ND and nPDF, see Table \ref{tab:tasks_overview}. These 3 were excluded from the benchmarking as they show a high degree of similarity with SAXS, XRD and xPDF respectively and the scattering intensity of X-rays is more predictable than neutrons as it scales with the number of electrons.

\textbf{Node-level tasks:} For all node-level tasks, the final GNN layer has feature dimensions matching the number of classes, see Table \ref{tab:tasks_overview}, or the size of the regression target, see Table \ref{tab:tasks_overview}.  

\textbf{Edge-level tasks:} For the edge-level task, the final GNN layer has 64 output features. The dot product of the output features of two nodes connected by an edge is then calculated to predict the distance associated with that edge.

\textbf{Graph-level tasks:} For all graph-level tasks, the final GNN layer has 64 output features. The graph is then summarized by concatenating a global mean, sum and max pooling, which is then passed through 2 fully connected linear layers with a ReLU activation function in between. The input size of the linear layers are 192, which is 3 times the number of output features, and the final output size matches the number of classes, see Table \ref{tab:tasks_overview}, or the size of the regression target, see Table \ref{tab:tasks_overview}. The hidden layer size is determined as the midpoint of the input and output dimensions.

\textbf{Generative tasks:} Before being parsed to the MLP model, the scattering data undergoes min-max normalization. The MLP model outputs a single vector of size equal to 3 times the maximum number of atoms (see Table \ref{tab:tasks_overview}) to accommodate all Cartesian coordinates ($x$, $y$ and $z$). A mask is derived from the ground truth structure to restrict the propagation of loss to errors on the coordinates of the atoms present in the ground truth structure. Before comparing the structures, the coordinates in the ground truth structure are sorted in ascending order according to their distance from the origin.

\section{Additional results}\label{app:additionalresults}

\subsection{Scattering data regression results}\label{app:scattering_regression}
\begin{figure}[!htb]
    \centering
    \includegraphics[width=\columnwidth]{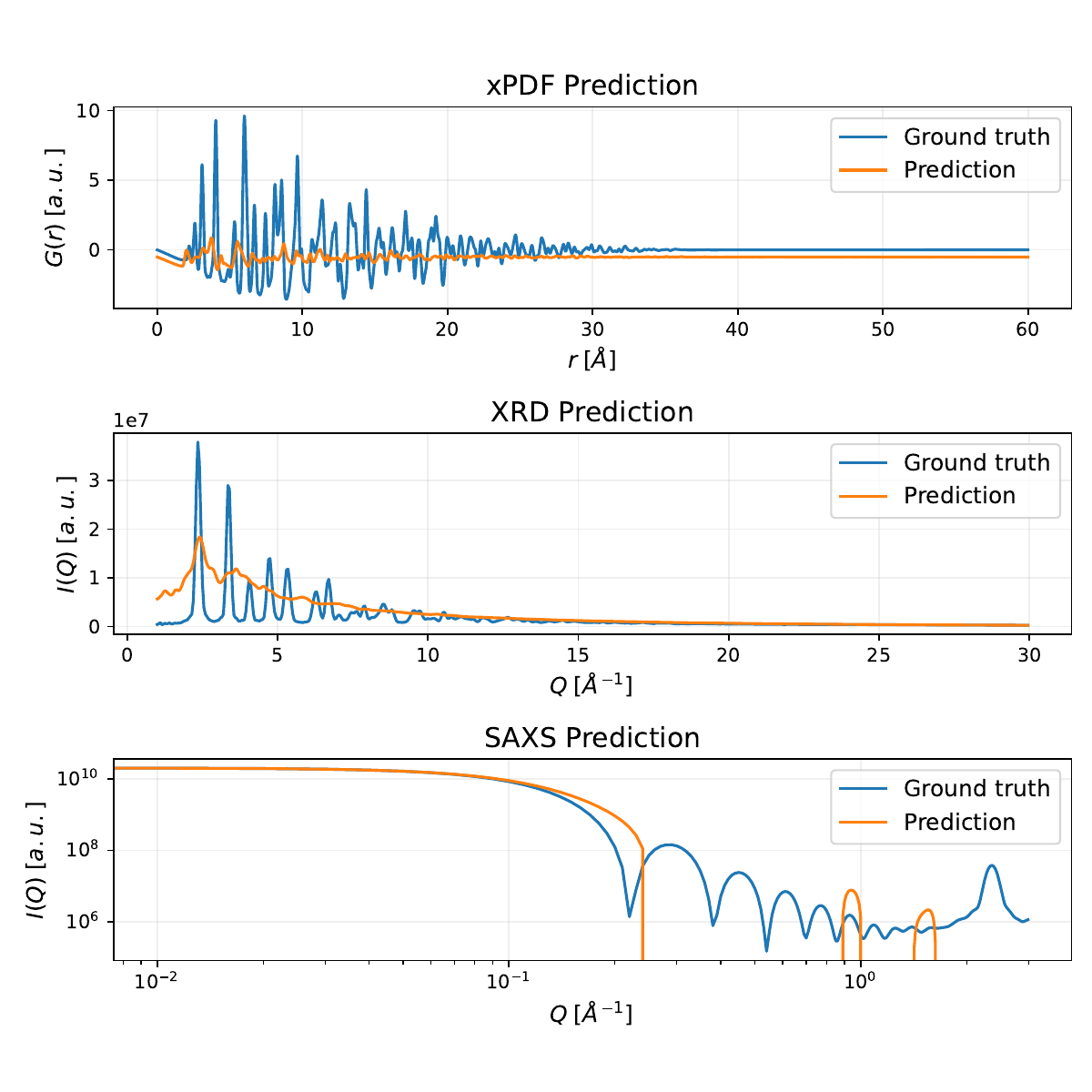}
    \vspace{-1.25cm}
    \caption{Regression results of the EdgeCNN model against the ground truth for the xPDF regression task (top), XRD regression task (middle), and SAXS regression task (bottom), all on a test sample of Mercury Oxide structure of crystal type NiAs from \texttt{CHILI-3K}. To improve clarity, both predictions and ground truths are rescaled to match the scale of the ground truth before normalization.}
    \label{fig:signal_regression_edgecnn_3k}
    \vspace{-0.25cm}
\end{figure}

\begin{figure}[!htb]
    \centering
    \includegraphics[width=\columnwidth]{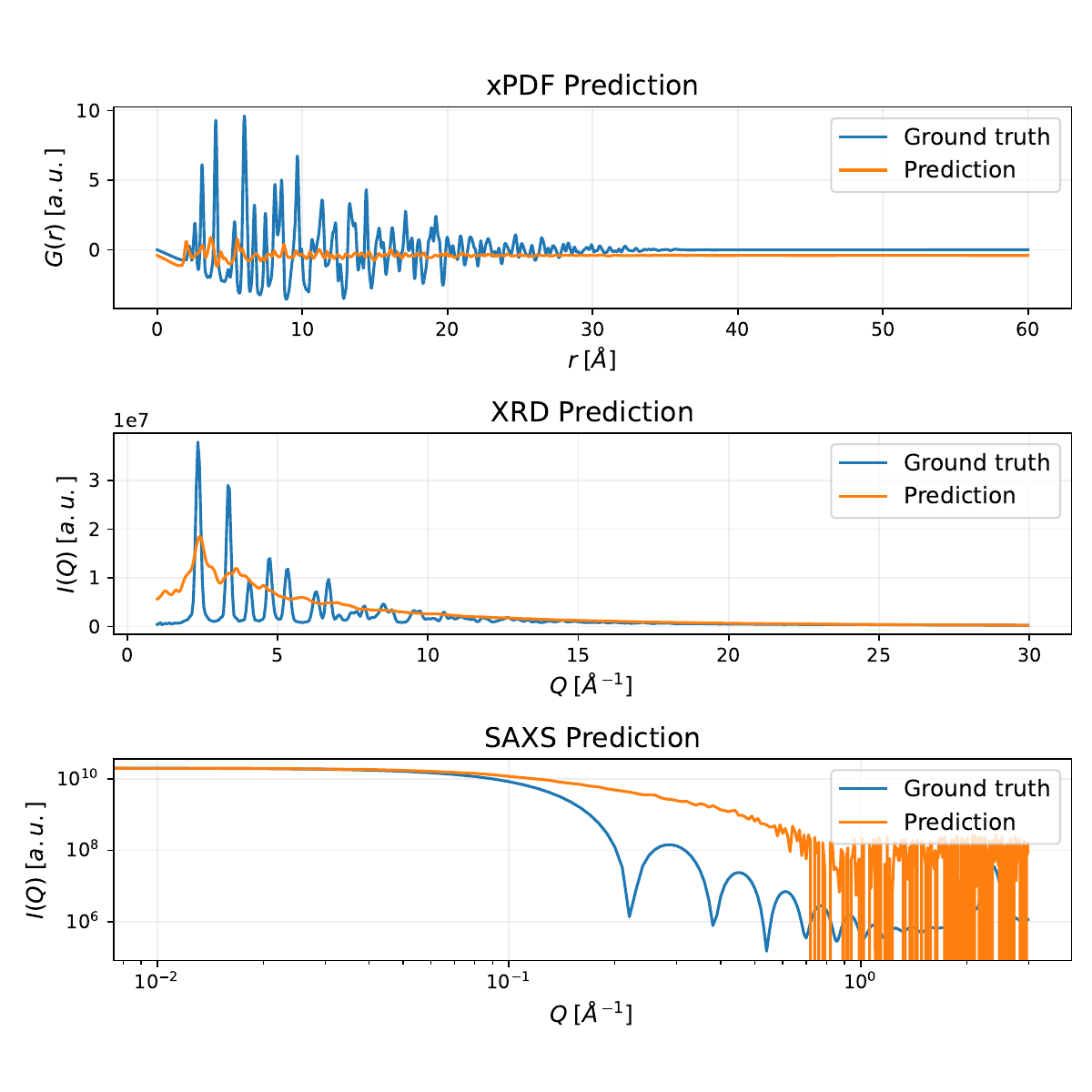}
    \vspace{-1.25cm}
    \caption{Regression results of the GAT model against the ground truth for the xPDF regression task (top), XRD regression task (middle), and SAXS regression task (bottom), all on a test sample of Mercury Oxide structure of crystal type NiAs from \texttt{CHILI-3K}. To improve clarity, both predictions and ground truths are rescaled to match the scale of the ground truth before normalization.}
    \label{fig:signal_regression_gat_3k}
    \vspace{-0.25cm}
\end{figure}

\begin{figure}[!htb]
    \centering
    \includegraphics[width=\columnwidth]{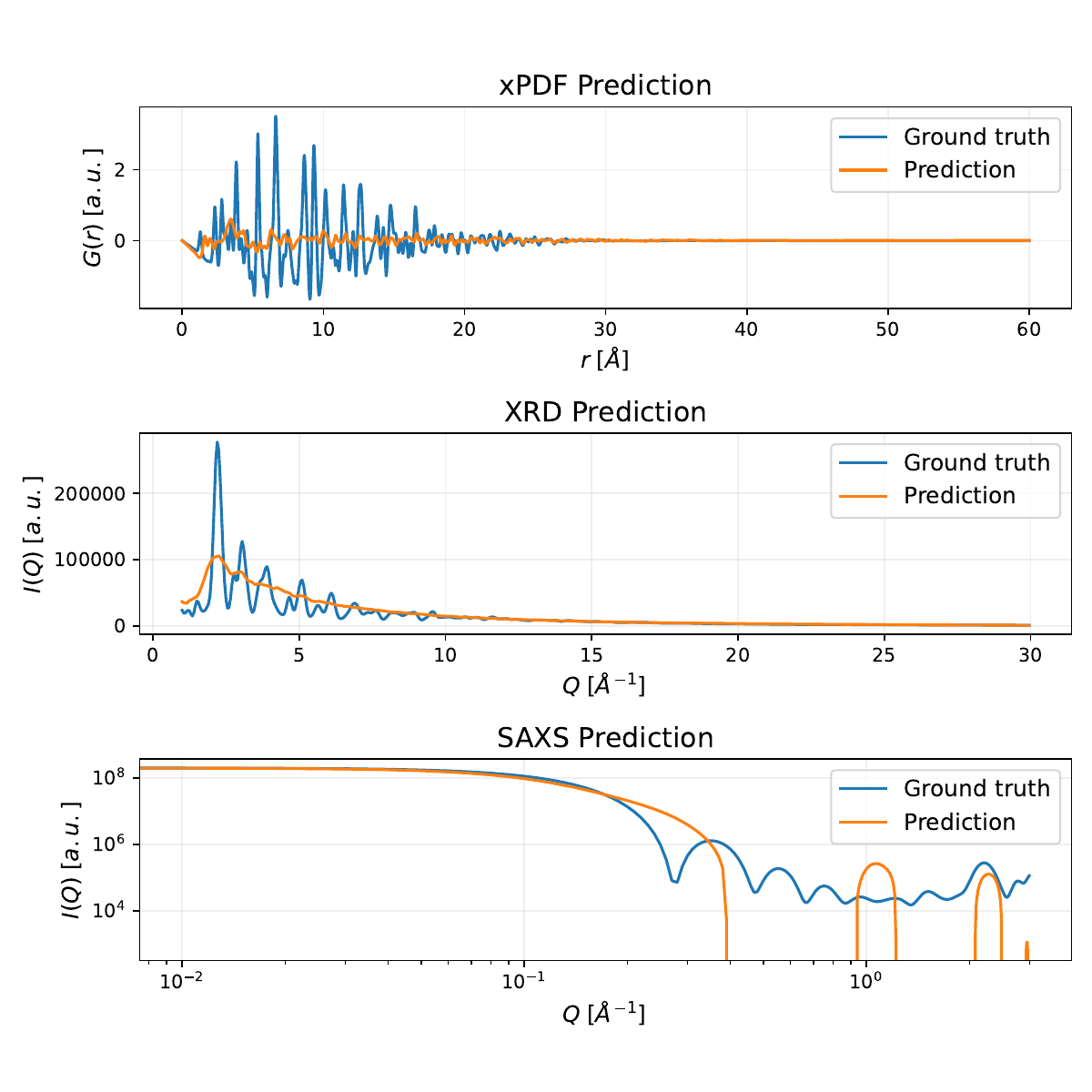}
    \vspace{-1.25cm}
    \caption{Regression results of the EdgeCNN model against the ground truth for the xPDF regression task (top), XRD regression task (middle), and SAXS regression task (bottom), all on a test sample of Butschliite (COD ID: 9000391) from \texttt{CHILI-100K}. To improve clarity, both predictions and ground truths are rescaled to match the scale of the ground truth before normalization.}
    \label{fig:signal_regression_edgecnn_100k}
    \vspace{-0.25cm}
\end{figure}

\begin{figure}[!htb]
    \centering
    \includegraphics[width=\columnwidth]{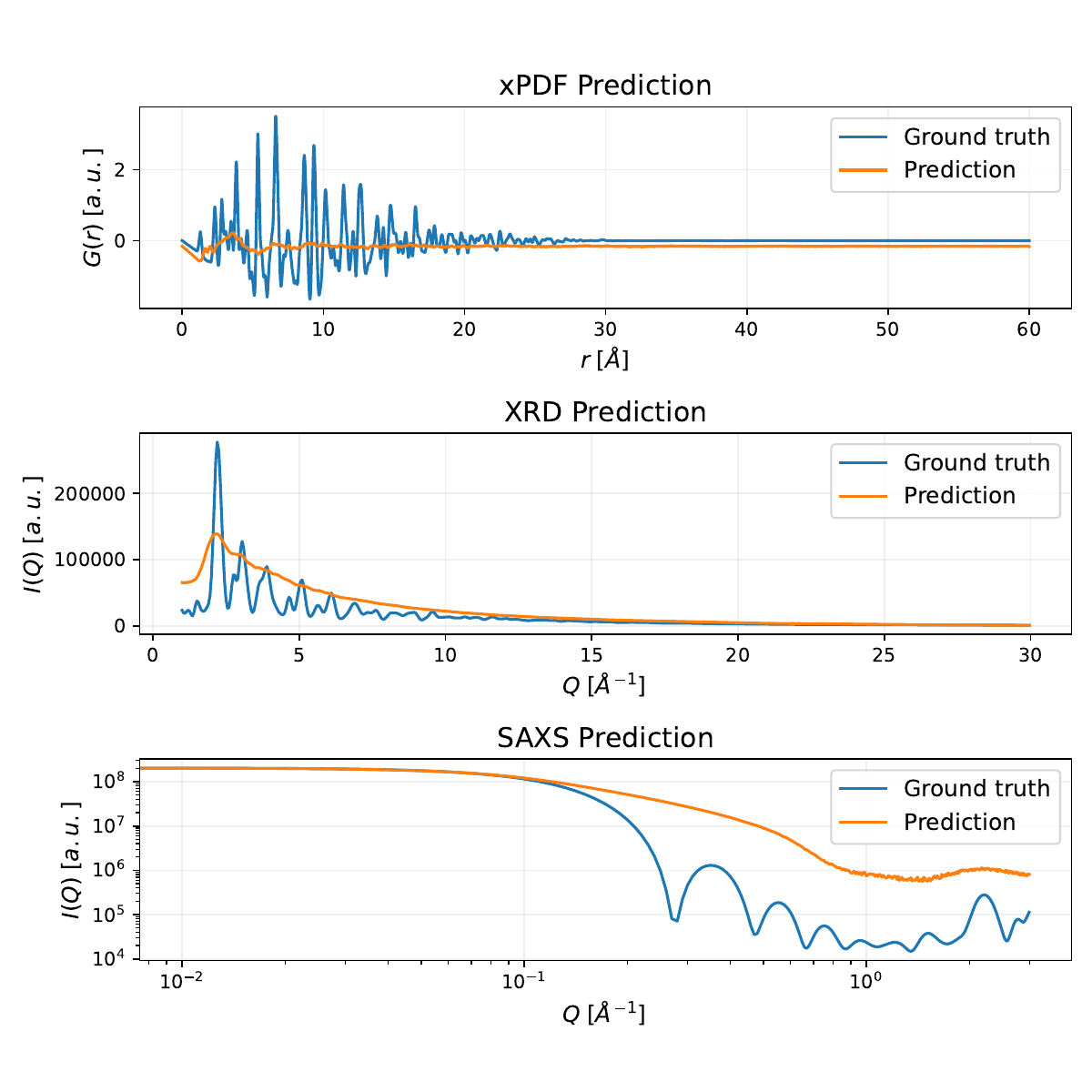}
    \vspace{-1.25cm}
    \caption{Regression results of the GAT model against the ground truth for the xPDF regression task (top), XRD regression task (middle), and SAXS regression task (bottom), all on a test sample of Butschliite (COD ID: 9000391) from \texttt{CHILI-100K}. To improve clarity, both predictions and ground truths are rescaled to match the scale of the ground truth before normalization.}
    \label{fig:signal_regression_gat_100k}
    \vspace{-0.25cm}
\end{figure}

We present regression results on test samples for predicting xPDF, XRD and SAXS; EdgeCNN on \texttt{CHILI-3K} in Figure \ref{fig:signal_regression_edgecnn_3k} and on \texttt{CHILI-100K} in Figure \ref{fig:signal_regression_edgecnn_100k}, GAT on \texttt{CHILI-3K} in Figure \ref{fig:signal_regression_gat_3k} and on \texttt{CHILI-100K} in Figure \ref{fig:signal_regression_gat_100k}

\subsection{Generative reconstruction results}\label{app:generative} 
In both the generative tasks, we first extract a subset of all structures within the datasets to train on; limiting them to a chosen maximum number of atoms. All structure coordinates are padded to ensure that their dimensionality matches and any loss calculations are restricted to operate solely on the atoms present in the ground truth structures. 

\begin{figure}[!htb]
    \centering

    \begin{subfigure}{0.3\columnwidth}
        \centering
        \includegraphics[width=\linewidth]{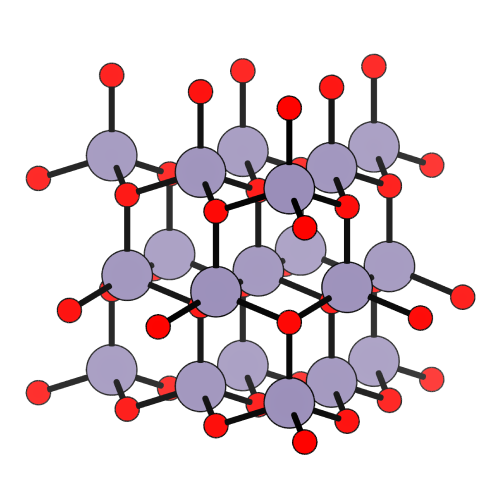}
        \caption{Ground truth}
        \label{fig:ground_truth_AbsPos}
    \end{subfigure}
    \begin{subfigure}{0.3\columnwidth}
        \centering
        \includegraphics[width=\linewidth]{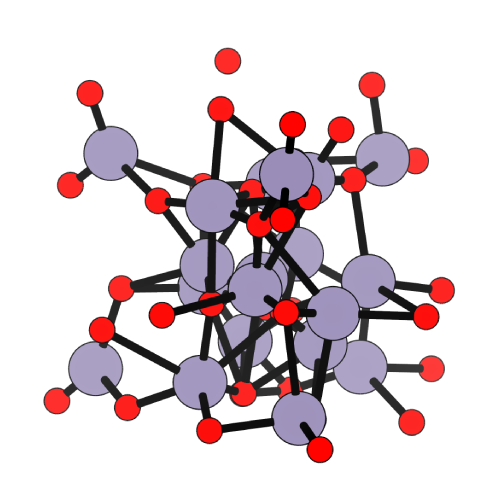}
        \caption{XRD}
        \label{fig:pred_xrd_AbsPos}
    \end{subfigure}
    \begin{subfigure}{0.3\columnwidth}
        \centering
        \includegraphics[width=\linewidth]{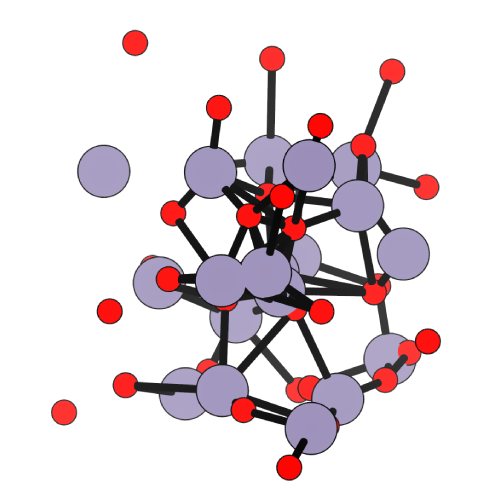}
        \caption{xPDF}
        \label{fig:pred_xPDF_AbsPos}
    \end{subfigure}
\vspace{-0.25cm}
    \caption{Reconstruction example using the MLP model on a sample from the \texttt{CHILI-3K} test set. The particle is derived from a Wurtzite crystal consisting of Tin (grey) and Oxygen (red). Panel (a) showcases the ground truth, while panels (b) and (c) display the predicted structures using XRD data (MAE: $2.19$ Å) and xPDF data (MAE: $2.48$ Å), respectively.}
    \label{fig:AbsReconstruction3K}
    \vspace{-0.5cm}
\end{figure}

\begin{figure}[!htb]
    \centering

    \begin{subfigure}{0.3\columnwidth}
        \centering
        \includegraphics[width=\linewidth]{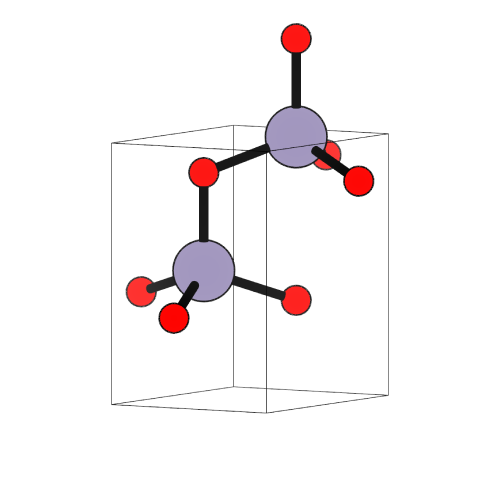}
        \caption{Ground truth}
        \label{fig:ground_truth_UnitPos_3K}
    \end{subfigure}
    \begin{subfigure}{0.3\columnwidth}
        \centering
        \includegraphics[width=\linewidth]{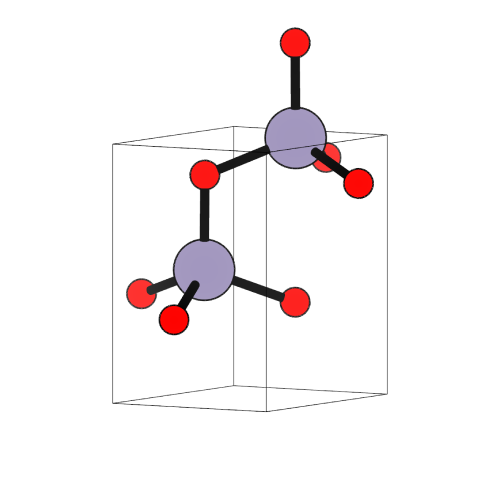}
        \caption{XRD}
        \label{fig:pred_xrd_UnitPos_3K}
    \end{subfigure}
    \begin{subfigure}{0.3\columnwidth}
        \centering
        \includegraphics[width=\linewidth]{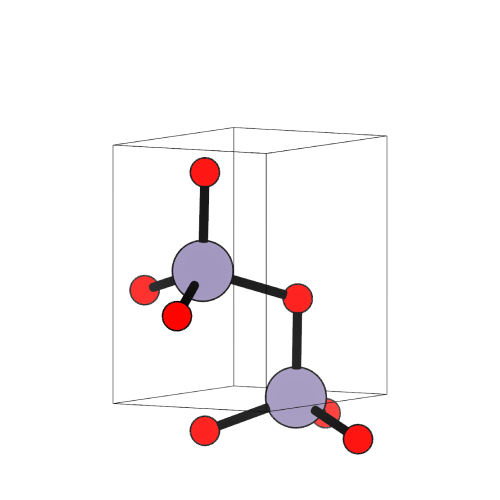}
        \caption{xPDF}
        \label{fig:pred_xPDF_UnitPos_3K}
    \end{subfigure}
\vspace{-0.15cm}
    \caption{Reconstruction example of unit cell using the MLP model on a sample from the \texttt{CHILI-3K} test set. The particle is derived from a Wurtzite crystal consisting of Tin (grey) and Oxygen (red). Panel (a) showcases the ground truth, while panels (b) and (c) display the predicted unit cell using XRD data (MAE: $0.0041$) and xPDF data (MAE: $0.0065$), respectively.}
    \label{fig:UnitReconstruction3K}
    \vspace{-0.25cm}
\end{figure}

\begin{figure}[!htb]
    \centering

    \begin{subfigure}{0.3\columnwidth}
        \centering
        \includegraphics[width=\linewidth]{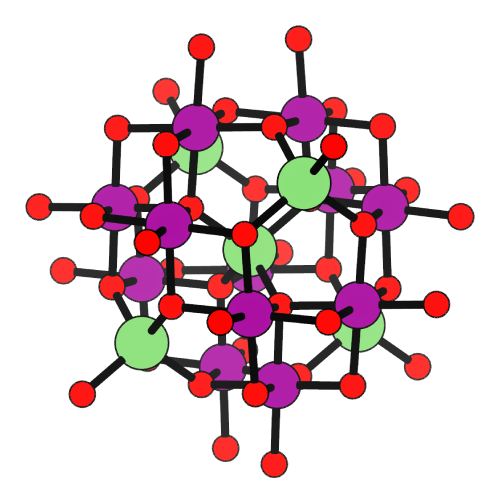}
        \caption{Ground truth}
        \label{fig:ground_truth_AbsPos_100K}
    \end{subfigure}
    \begin{subfigure}{0.3\columnwidth}
        \centering
        \includegraphics[width=\linewidth]{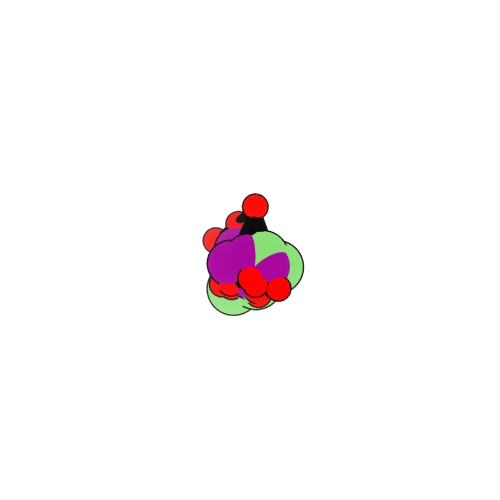}
        \caption{XRD}
        \label{fig:pred_xrd_AbsPos_100K}
    \end{subfigure}
    \begin{subfigure}{0.3\columnwidth}
        \centering
        \includegraphics[width=\linewidth]{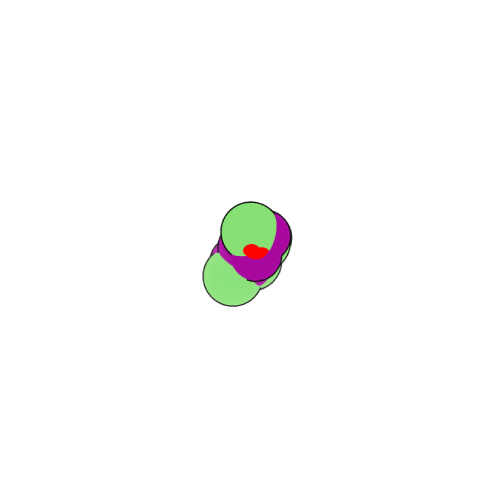}
        \caption{xPDF}
        \label{fig:pred_xPDF_AbsPos_100K}
    \end{subfigure}
\vspace{-0.25cm}
    \caption{Reconstruction example using the MLP model on a sample from the \texttt{CHILI-100K} test set. The particle is derived from a Lithium Manganese Oxide crystal (COD-ID: 7214197) consisting of Lihium (green), Maganese (purple) and Oxygen (red). Panel (a) showcases the ground truth, while panels (b) and (c) display the predicted structures using XRD data (MAE: $1.84$ Å) and xPDF data (MAE: $1.81$ Å), respectively.}
    \label{fig:AbsReconstruction100K}
\end{figure}

\begin{figure}[!htb]
    \centering

    \begin{subfigure}{0.3\columnwidth}
        \centering
        \includegraphics[width=\linewidth]{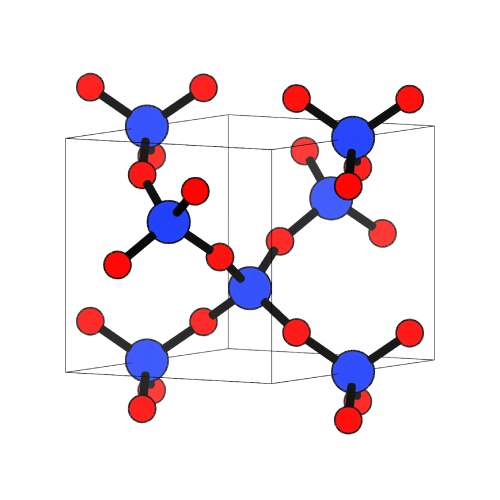}
        \caption{Ground truth}
        \label{fig:ground_truth_UnitPos_100K}
    \end{subfigure}
    \begin{subfigure}{0.3\columnwidth}
        \centering
        \includegraphics[width=\linewidth]{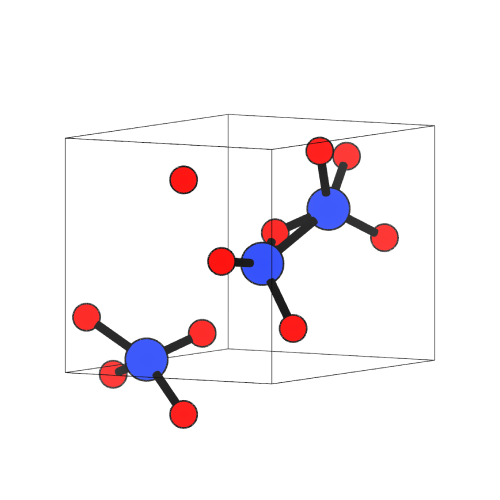}
        \caption{XRD}
        \label{fig:pred_xrd_UnitPos_100K}
    \end{subfigure}
    \begin{subfigure}{0.3\columnwidth}
        \centering
        \includegraphics[width=\linewidth]{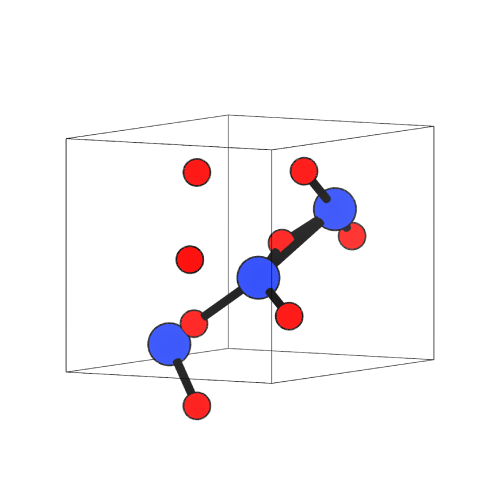}
        \caption{xPDF}
        \label{fig:pred_xPDF_UnitPos_100K}
    \end{subfigure}

    \caption{Reconstruction example of unit cell using the MLP model on a sample from the \texttt{CHILI-100K} test set. The particle is derived from a \ce{SiO2} Quartz crystal consisting of Silicon (blue) and Oxygen (red). Panel (a) showcases the ground truth, while panels (b) and (c) display the predicted unit cell using XRD data (MAE: $0.138$) and xPDF data (MAE: $0.151$), respectively.}
    \label{fig:UnitReconstruction100K}
\end{figure}

Figure \ref{fig:AbsReconstruction3K} shows a reconstruction example using the MLP model on a sample from \texttt{CHILI-3K}, predicting absolute positions. Figure \ref{fig:AbsReconstruction100K} shows a reconstruction example using the MLP model on a sample from \texttt{CHILI-100K}, predicting absolute positions. Figure \ref{fig:UnitReconstruction3K} and \ref{fig:UnitReconstruction100K} each show equivalent examples, but for the unit cells.

\subsection{Property-to-property prediction results}\label{app:p2p}
In addition to the standard classification, regression and generation tasks, we performed benchmarks on tasks which involve the classification of crystal systems and spacegroups from scattering data, as well as regression of the crystal unit cell parameters. The results of these benchmarks are presented in Table \ref{tab:results_p2p}.

\begin{table}[!htb]
    \caption{Results on property-to-property benchmark experiments on the \texttt{CHILI-3K} and \texttt{CHILI-100K} datasets. Metrics are: F1-score for all classification tasks and MSE for all regression tasks.}
    \label{tab:results_p2p}
    \small
    \begin{tabular}{lcc}
        \toprule
        Task & \texttt{CHILI-3K} & \texttt{CHILI-100K} \\
        \midrule
        CrystalSystemClassificationSAXS & $0.849 \pm 0.049$ & $0.253 \pm 0.012$ \\
        CrystalSystemClassificationXRD & $0.983 \pm 0.006$ & $0.475 \pm 0.009$ \\
        CrystalSystemClassificationxPDF & $0.975 \pm 0.006$ & $0.185 \pm 0.100$ \\
        \midrule
        SpacegroupClassificationSAXS & $0.828 \pm 0.022$ & $0.146 \pm 0.014$ \\
        SpacegroupClassificationXRD & $0.983 \pm 0.008$ & $0.364 \pm 0.015$ \\
        SpacegroupClassificationxPDF & $0.958 \pm 0.004$ & $0.314 \pm 0.039$ \\
        \midrule
        CellParamsRegressionSAXS & $42.726 \pm 10.425$ & $73.501 \pm 1.214$ \\
        CellParamsRegressionXRD & $3.533 \pm 1.575$ & $62.621 \pm 7.219$ \\
        CellParamsRegressionxPDF & $3.554 \pm 1.081$ & $70.355 \pm 2.939$ \\
        \bottomrule
    \end{tabular}
\end{table}

\clearpage
\end{document}